\documentclass{article}

\usepackage{PRIMEarxiv}

\usepackage[utf8]{inputenc} 
\usepackage[T1]{fontenc}    
\usepackage[hidelinks]{hyperref}       
\usepackage{url}            
\usepackage{booktabs}       
\usepackage{amsfonts}       
\usepackage{nicefrac}       
\usepackage{microtype}      
\usepackage{lipsum}
\usepackage{fancyhdr}       
\usepackage{graphicx}       
\usepackage{multirow}
\usepackage[numbers,sort&compress]{natbib}
\usepackage{algorithm}
\usepackage{algpseudocode}
\usepackage{amsmath}
\graphicspath{{media/}}     

\usepackage{subfigure}

\title{LLM-Based SQL Generation: Prompting, Self-Refinement, and Adaptive Weighted Majority Voting
}

\author{
  Yu-Jie Yang\\
  Institute of Information Management \\
  National Yang Ming Chiao Tung University \\
  \texttt{estheryangyujie@nycu.edu.tw} \\
   \And
  Hung-Fu Chang  \\
  R. B. Annis School of Engineering \\
  University of Indianapolis \\
  \texttt{hchang@uindy.edu} \\
  \And
   Po-An Chen \\
   Institute of Information Management \\
   National Yang Ming Chiao Tung University \\
   \texttt{poanchen@nycu.edu.tw} \\
}

\begin{document}
\maketitle

\begin{abstract}
Text-to-SQL has emerged as a prominent research area, particularly with the rapid advancement of large language models (LLMs). By enabling users to query databases through natural language rather than SQL, this technology significantly lowers the barrier to data analysis. However, generating accurate SQL from natural language remains challenging due to ambiguity in user queries, the complexity of schema linking, limited generalization across SQL dialects, and the need for domain-specific understanding. In this study, we propose a Single-Agent Self-Refinement with Ensemble Voting (SSEV) pipeline built on PET-SQL that operates without ground-truth data, integrating self-refinement with Weighted Majority Voting (WMV) and its randomized variant (RWMA). Experimental results show that the SSEV achieves competitive performance across multiple benchmarks, attaining execution accuracies of 85.5\% on Spider 1.0-Dev, 86.4\% on Spider 1.0-Test, and 66.3\% on BIRD-Dev. Building on insights from the SSEV pipeline, we further propose ReCAPAgent-SQL (Refinement-Critique-Act-Plan agent-based SQL framework) to address the growing complexity of enterprise databases and real-world Text-to-SQL tasks. The framework integrates multiple specialized agents for planning, external knowledge retrieval, critique, action generation, self-refinement, schema linking, and result validation, enabling iterative refinement of SQL predictions through agent collaboration. ReCAPAgent-SQL's WMA results achieve 31\% execution accuracy on the first 100 queries of Spider 2.0-Lite, demonstrating significant improvements in handling real-world enterprise scenarios. Overall, our work facilitates the deployment of scalable Text-to-SQL systems in practical settings, supporting better data-driven decision-making at lower cost and with greater efficiency.
\end{abstract}

\keywords{Text-to-SQL, Large Language Models (LLMs), Weighted Majority Algorithm (WMA), Self-Refinement}

\section{Introduction}
Text-to-SQL has long been a core research problem at the intersection of natural language processing and database systems. In industrial settings, data scientists and data analysts are often required to translate business analytics questions expressed in natural language into efficient SQL queries over complex database schemas and large-scale datasets. To enable convenience, companies such as Salesforce and Snowflake have integrated Text-to-SQL capabilities into their Business Intelligence (BI) platforms \cite{WrenAI2024GenBI}, reducing reliance on specialized SQL expertise and mitigating the data access gap\cite{salesforce_text2sql,snowflake_aisql}. Despite increasing adoption and encouraging practical results, existing systems continue to face substantial challenges when deployed in complex real-world scenarios \cite{lei-2024}. These limitations stem in part from early Text-to-SQL approaches, which relied heavily on manual scripting, templates, or rule-based methods that required extensive domain expertise and scaled poorly across heterogeneous databases.
 
Recently, the rapid advancement of large language models (LLMs), such as GPT, Gemini, Grok, Qwen, and Llama, has significantly reshaped the Text-to-SQL landscape. These models have demonstrated strong capabilities in natural language understanding and code generation, spurring growing interest in applying LLMs to software development and database management tasks. Prior LLM-based Text-to-SQL studies, such as Spider 1.0 \cite{yu-2018-spider} and BIRD \cite{li2023bird}, have achieved strong performance on benchmark datasets; however, several critical challenges remain unresolved. In particular, existing approaches often fail to effectively integrate external knowledge, contextual information, and step-by-step reasoning \cite{deng2025reforce}. Schema linking is typically limited to partial schema information, resulting in incomplete reasoning over implicit joins and cross-table references. Moreover, generalization across diverse SQL dialects and query styles remains weak, as evidenced by frequent failures on realistic datasets such as Spider 2.0 \cite{lei-2024}. Finally, most Text-to-SQL pipelines rely on a single model to generate the final SQL output, lacking robust output selection strategies, adaptive feedback mechanisms, and the ability to learn from prior experience.

To address these deficiencies, our approach centers on leveraging multiple LLMs as complementary experts, combined with multi-agent reasoning, to resolve ambiguity, enhance schema understanding, improve generalization, and satisfy deployment constraints. To coordinate predictions from multiple LLM experts without requiring ground-truth SQL at inference time, we suggest the Weighted Majority Algorithm (WMA) and its randomized variant, the Randomized Weighted Majority Algorithm (RWMA). By dynamically updating expert weights based on historical performance, the framework progressively prioritizes more reliable experts while maintaining constant-time online inference and relegating computationally intensive updates to offline processing. To systematically evaluate the effectiveness of these ideas, we adopt a phased evaluation strategy. In the first phase, we develop a Single-Agent Self-Refinement with Ensemble Voting (SSEV) pipeline and evaluate it on relatively simpler benchmarks, Spider 1.0 and BIRD. Building on the insights gained from the first phase, we then extend the single-agent approach to a multi-agent setting to create the ReCAPAgent-SQL (Refinement–Critique–Act–Plan Agent-based SQL) framework, which is evaluated on the more challenging Spider 2.0 dataset.

Results from this two-phase evaluation demonstrate that our design choices effectively mitigate the challenges outlined above. In the first stage, our single-agent framework achieves execution accuracies of up to 86.4\% on Spider 1.0 and 66.3\% on BIRD, outperforming strong individual LLM baselines while preserving O(1) online latency. In the second stage, ReCAPAgent-SQL improves execution accuracy on Spider 2.0-lite from 6\% to 31\% using GPT-4.1 and from 23\% to 29\% using Grok-3-beta. Incorporating WMA and RWMA preserves these performance gains, demonstrating robustness and adaptability under limited supervision.

\section{Related Work}
\subsection{LLM-based Text-to-SQL Overview}

Most early Text-to-SQL approaches relied on rule-based methods. With the development of machine learning, encoder-encoder models like Seq2Seq, SQLNet, and TypeSQL \cite{yu-2018-spider} were applied to extracting the database information. Text-to-SQL systems convert natural language queries into SQL statements, enabling more direct database access for non-technical users. In recent years, LLMs (e.g., GPT, LLaMa) have significantly developed by enabling in-context learning and prompt-based query generation without fine-tuning. Text-to-SQL workflow approaches like DIN-SQL \cite{Pourreza2023din}, DAIL-SQL \cite{gao2023texttosql}, and PET-SQL \cite{li2024pet} have leveraged LLMs to provide high accuracy on cross-domain datasets such as Spider 1.0 and BIRD. In Next-Generation Database Interfaces: A Survey of LLM-based Text-to-SQL \cite{Hong2024NextGen}, they present a comprehensive analysis of current LLM-based Text-to-SQL challenges to be addressed, such as linguistic complexity, semantic ambiguity, schema understanding, complex SQL operations, and cross-domain generalization. Furthermore, in real-world enterprise scenarios, data workflows are so complicated that Text-to-SQL systems need a multi-step reasoning plan and iterative steps to address SQL queries \cite{lei-2024}.

\subsection{Prompting Strategies and Two-Stage SQL Generation}
Prompting-based methods have become a classic method for adapting LLMs to Text-to-SQL tasks by taking advantage of in-context learning with a set of Question-SQL examples as demonstrations, i.e., few-shot prompts. Furthermore, exploring the idea of prompt-based SQL generation, the study, \textit{``Text-to-SQL Empowered by Large Language Models: A Benchmark Evaluation``} \cite{gao2023texttosql}, systematically compared all existing prompt engineering methods, such as question representation, example selection, and example organization, for their effects on Text-to-SQL performance. From those findings, Gao et al. \cite{gao2023texttosql} proposed an integrated solution called \textsc{DAIL-SQL}, they featured the example selection (\textsc{DAIL}\textsubscript{s}) with few shots and example organization (\textsc{DAIL}\textsubscript{O}) to achieve 86.6\% execution accuracy on Spider 1.0 \cite{gao2023texttosql}.
Viewing DAIL-SQL \cite{gao2023texttosql} as the foundation, Liu et al.~(2024) \cite{li2024pet} put forth \textsc{PET-SQL}: a prompt-enhanced two-round refinement framework. They proposed the Reference-Enhanced representation (\textsc{RE}\textsubscript{p}) based on OpenAI Demonstration (\textsc{OD}\textsubscript{p}) that recommended in \cite{gao2023texttosql}. Therefore, in the first stage, SQL was generated preliminarily (PreSQL) from a few-shot demonstration; in the second stage, schema linking was used to further update the prompt's schema information for PostSQL generation. They reported execution accuracy as high as 87.6\% on the Spider 1.0 benchmark. Prompting methods truly enhanced the Text-to-SQL accuracy.

\subsection{Self-Consistency and Cross-Consistency}

Self-consistency has been introduced in previous work, such as DAIL-SQL \cite{gao2023texttosql}. The goal is to generate multiple SQL predictions from a single LLM model at a higher temperature and select the final answer by majority voting. Although self-consistency can enhance output stability, it also often leads to hallucinations or invalid queries due to semantic errors since sampling at higher temperatures increases randomness instead of improving diversity.

However, to overcome these shortcomings, PET-SQL \cite{li2024pet} proposed an alternative strategy called Cross-Consistency. Instead of relying on the single LLM model, the approach implements multiple LLMs (e.g., GPT-4o, CodeLlama, SQLCoder) to generate predicted SQL statements at a low temperature. Then, they used the Naïve or difficulty-aware approach to vote, improving both semantic correctness and adaptability across SQL complexities.

PET-SQL \cite{li2024pet} demonstrates that \textit{cross-consistency} outperforms \textit{self-consistency} by a large gain. On the Spider 1.0 benchmark, although self-consistency yields marginal improvements (e.g., +0.8\%), cross-consistency improves execution accuracy by up to +6.9\%, achieving 87.6\% execution accuracy (EX) on the Spider 1.0. Additionally, it also shows that \textit{difficulty-aware voting} not only enhances performance by selecting specific models based on the difficulty of the gold query but also outperforms single-model strategies in Spider 1.0 evaluation \cite{yu-2018-spider}.

\subsection{Agent-based Reasoning in Text-to-SQL Systems}

As Text-to-SQL applications were broadly implemented in real-world enterprise environments, the limitations of static single-turn LLM prompting became increasingly evident. To address these challenges, recent research proposed agent-based reasoning frameworks that enabled LLMs to interact dynamically with database environments, external documents, and execution feedback in a multi-turn manner.

 \cite{lei-2024} not only proposed Spider 2.0 but also pointed out the need for autonomous agents capable of planning, reasoning, and optimizing SQL queries across different SQL dialects. While traditional Text-to-SQL frameworks performed well on Spider 1.0 and BIRD, they still performed poorly on Spider 2.0 tasks \cite{lei-2024}, achieving no higher than 10\% execution accuracy. Faced with such difficulties, the authors introduced Spider-Agent, a ReAct-style agent framework that iteratively interacted with databases and codebases, demonstrating the advantages of planning and multi-stage execution.

\subsection{Retrieval-Augmented Generation (RAG) for Text-to-SQL and Knowledge Grounding}
Retrieval-Augmented Generation (RAG) has been considered a powerful method for augmenting LLMs with searching external, task-specific knowledge, especially when domain-specific information is not fully ingested within the model’s pretraining corpus. In the field of Text-to-SQL, RAG enables dynamic access to relevant documentation, syntax references, and domain-specific knowledge, which is vitally necessary for accurate SQL generation in enterprise environments involving complex dialect functions like BigQuery or Snowflake.

Previous work, such as DB-GPT \cite{Xue2023DBGPT}, highlights the practical potential of RAG in database interaction scenarios. DB-GPT takes advantage of a multi-source knowledge base, combining documents, web pages, and structured content, to support context-query generation. Their architecture includes three main phases: (1) knowledge construction via paragraph-level embedding, (2) retrieval via similarity search using multilingual embedding models, and (3) adaptive in-context learning (ICL) for response generation. This design enables DB-GPT to provide accurate SQL generation, question answering, and bilingual interaction while maintaining data privacy in local deployments.

\subsection{Self-Refinement in Text-to-SQL Systems}
Self-refinement mechanisms serve as a critical post-generation correction technique in modern Text-to-SQL systems. These techniques have an iterative query revision based on execution feedback and error diagnosis, significantly enhancing SQL robustness in both benchmark and real-world scenarios.

Building on these insights and Spider-Agent \cite{lei-2024}, ReFoRCE \cite{deng2025reforce} introduces a self-refinement framework that is able to categorize SQL execution errors (e.g., \texttt{ColumnNotFound}, \texttt{SyntaxError}, \texttt{AmbiguousColumn}) and applies tailored repair strategies. It implements techniques such as Common Table Expression (CTE)-based rewriting, column probing, and format enforcement. Moreover, the refinement process is iterated until termination conditions such as fixed output, maximum attempts, or consistent failure are met.

Overall, ReFoRCE showcased its improvements on real-world datasets like Spider 2.0-lite and Spider 2.0-snow through structured feedback in its refinement process. On May 22\textsuperscript{nd}, 2025, ReFoRCE achieves 37\% on Spider 2.0-lite with the o3 model and demonstrates that advanced self-refinement plays a pivotal role in complex enterprise applications.

\section{Methodology}
\label{ch:method}
\subsection{Common Techniques in SSEV pipeline and ReCAPAgent-SQL}
Inspired by PET-SQL \cite{li2024pet}, a prompt-enhanced two-round Text-to-SQL framework that achieves strong execution accuracy on Spider 1.0, our approach firstly incorporates PreSQL generation, schema linking, and PostSQL generation, together with carefully designed prompts and example selection. Then, building on these components, we integrate execution-guided self-refinement and adaptive expert voting to enhance the overall pipeline. For expert voting, we employ the Weighted Majority Algorithm (WMA) and its randomized variant (RWMA). In the following sections, we describe these techniques in detail, as they are shared by both of our proposed approaches.

\subsubsection{PreSQL, Schema Linking, and PostSQL}
\label{sec:schema_linking}
Three major components are included in the two-round PET-SQL method. In the first round, PreSQL generation is used to construct a SQL-tailored prompt. This prompt is then combined with schema linking, where the linking process is guided by the generated PreSQL. Based on the preliminary SQL produced in the first round, a schema-informed prompt is constructed for the second round, referred to as PostSQL generation. By leveraging the rough SQL from the first round, the model can reduce irrelevant token noise and restrict attention to a smaller set of relevant tables and columns in the PostSQL round, thereby improving the accuracy of the final SQL generation.

\begin{itemize}
    \item \textbf{PreSQL Generation:} \\
    PET-SQL deploys N different LLMs (e.g., InterLM-70B, SQLCoder-34B, CodeLlama-34B) in parallel. Each LLM is configured with identical hyperparameters (a low temperature of approximately 0.0 and a maximum token limit of 200) and generates a single PreSQL candidate.

    Several key components are considered to ensure effective prompt design and example selection during the PreSQL round for prompt construction. They determine both the structure of the prompt and the contextual information.

    \begin{itemize}
      \item \textbf{Task Instruction \& Optimization Rule:} 
        A strict instruction to minimize SQL execution time is included in the prompt. This constraint guides the model toward generating efficient SQL queries while avoiding unnecessary complexity and computationally expensive operations.
    
        \begin{figure}[htb]
          \centering
          \includegraphics[width=0.6\textwidth]{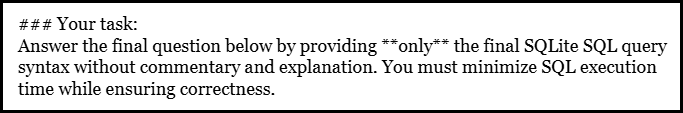}
          \caption{Task Instruction \& Optimization Rule Prompt Example.}
          \label{fig:TaskInstruction}
        \end{figure}
        
      \item \textbf{Full Database DDL \& Schema Content:}  
        The prompt includes a complete description of the database schema, consisting of all table names, column names, and foreign-key relationships, presented in standard SQL DDL format. An example of the SQL DDL format is shown in Figure 2.
        
        \begin{figure}[htb]
          \centering
          \includegraphics[width=0.6\textwidth]{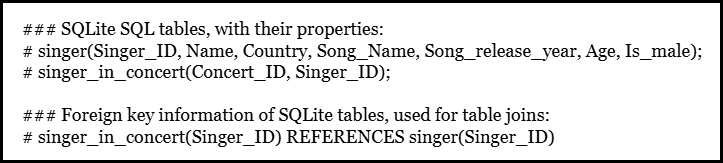}
          \caption{Full Database DDL \& Optimization Rule Prompt Example.}
          \label{fig:Full Database DDL Schema}
        \end{figure}
        Including these DDL comments allows the LLM to “see” the schema information.
    
      \item \textbf{Cell Value References:}  
        The prompt includes randomly sampled representative rows (three per table). These cell-level references help resolve formatting ambiguities (e.g., “Male” vs. “M”).
       
        \begin{figure}[htb]
          \centering
          \includegraphics[width=0.6\textwidth]{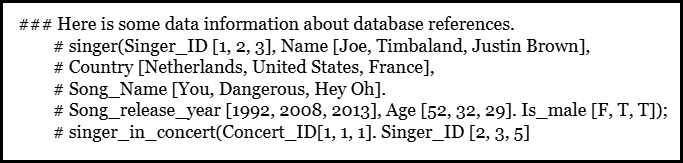}
          \caption{Cell Value References Prompt Example.}
          \label{fig: Cell Value References}
        \end{figure}
            
      \item \textbf{Few‐Shot Demonstrations:}  
        Based on embedding similarity, the top-k question–SQL pairs most similar to the input query are retrieved. These examples guide the model toward generating SQL that is stylistically and structurally consistent with the target dataset.
        
        \begin{figure}[htb]
          \centering
          \includegraphics[width=0.35\textwidth]{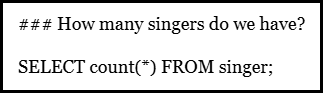}
          \caption{Few‐Shot Demonstrations Prompt Example.}
          \label{fig:Few Shot Demonstrations}
        \end{figure}
    \end{itemize} 

    After assembling the complete prompt with optimization rule, full schema DDL, sampled cell value references, and k few-shot examples, a prompt is submitted to an LLM to generate the first-round SQL, referred to as PreSQL.
    
    \item \textbf{PreSQL-Based Schema Linking:} \\
    Schema linking is considered a long-standing and fundamentally crucial component for Text-to-SQL since it bridges natural language utterances with structured database schemas. During preprocessing, the mappings are identified between query tokens (e.g., column names, values) and schema elements (e.g., tables, attributes). With the emergence of large language models (LLMs), schema linking has evolved from explicit rule-based modules to implicit attention-based mechanisms, leveraging pre-trained embeddings and prompt-based cues. Recent studies further treat schema linking as an independent and controllable factor influencing model performance. For example, Gao et al.  \cite{gao2023texttosql} evaluated multiple schema linking strategies and demonstrates their substantial impact on downstream SQL generation quality.

    The PET-SQL prompt initially includes the full database schema, consisting of all tables and columns. However, each PreSQL candidate typically references only a subset of this schema. Each PreSQL string is parsed to extract fully qualified table and column names (e.g., using regular expressions for patterns such as <table>.<column> or FROM <table>). The union of all extracted schema elements across the N PreSQL candidates is then treated as the linked schema, representing the minimal set of tables and columns deemed relevant by the LLMs. All other schema elements are pruned from the subsequent prompt.

    \item \textbf{PostSQL (i.e., final SQL) Generation:} \\
    A concise, schema-informed prompt starts to be constructed by the same template in the PreSQL while replacing the full database DDL with a pruned DDL containing only the related tables and columns identified in the PreSQL round. Each LLM then generated a PostSQL candidate with this pruned prompt and identical settings.
\end{itemize}

\subsubsection{Weighted Majority Algorithm Voting Mechanisms}

The Weighted Majority Algorithm (WMA), initially introduced by Littlestone and Warmuth \cite{shalevCMU15850}, is a foundational ensemble technique in online learning that combines predictions from multiple experts.While  WMA and its randomized (RWMA) \cite{hazan2023oco} have been widely used in online decision-making and adversarial environments, their application to the Text-to-SQL domain remains novel.
\begin{itemize}
    \item \textbf{Weighted Majority Algorithm (WMA):} It assigns a fixed weight for each expert and penalizes incorrect predictions by decreasing the expert's weight. Over time, better-performing experts maintain the higher weights, allowing the system to gradually favor the most reliable ones.
    \item \textbf{Randomized Weighted Majority Algorithm (RWMA) : } It is a probabilistic variant of WMA, sampling experts according to their weights instead of taking a deterministic maximum. This adds robustness under adversarial or noisy environments by avoiding overfitting to a single model.
\end{itemize}

In both our SSEV pipeline and ReCAPAgent-SQL, WMA adaptively weights each expert (i.e., LLM) based on its historical performance, allowing the stronger model to have a greater influence over the final prediction. Initially, each expert $i \in \{1, \dots, N\}$ is assigned a uniform weight:

\begin{equation}
    w_i = 1.0
\end{equation}

Given a set of SQL candidates $\{s\}$ proposed by different experts, the total weight for each SQL $s$ is computed as:

\begin{equation}
    W(s) = \sum_{i : s \in \mathcal{S}_i} w_i
\end{equation}

The SQL with the highest total weight is selected as the final output. After each iteration, experts' weights are updated based on correctness (if gold labels are available) or execution success. Incorrect outputs lead to multiplicative penalties by a factor of $(1-\epsilon)$. This strategy is formally detailed in the following: 

\begin{algorithm}
\caption{Weighted Majority Voting Algorithm}
\begin{enumerate}
\item Initialize weights: $w_i = 1.0$ for each expert $i$.
  
  \item Collect predictions: $\mathcal{S}_i \leftarrow$ SQL candidates from expert $i$.
  
  \item Aggregate: For each unique SQL $s$, compute the weight sum
  \[
    W(s) = \sum_{i : s \in \mathcal{S}_i} w_i.
  \]

  \item Select: $\hat{s} = \arg\max_s W(s)$.

  \item Update (if gold SQL available): If expert $i$'s output was incorrect,
  \[
    w_i \leftarrow w_i \cdot (1 - \varepsilon).
  \]
  
  \item \textbf{Epsilon setting:} To minimize the mistake bound, set
  \[
    \varepsilon = \sqrt{\frac{\log N}{T}},
  \]

  \item \textbf{Mistake Bound:}
  \[
    M_T \leq M^*(2 + \varepsilon) + \frac{2 \log N}{\varepsilon} \leq 2M^* + 4\sqrt{T\log N}
  \]
    where $N$ is the number of experts, and $M^*$ is the number of mistakes made by the best expert during $T$ rounds.
\end{enumerate}
\end{algorithm}

We further extend our method with the \textit{Randomized Weighted Majority Algorithm (RWMA)}. Similarly, RWMA utilizes the same update rule as WMA, but it samples the final SQL probabilistically according to the normalized weights of the contributing experts. This randomized decision-making process introduces diversity in model selection. It ensures that even lower-weighted experts can influence the final decision. The complete RWMA procedure is outlined  as follows: 

\begin{algorithm}[H]
\caption{Randomized Weighted Majority Voting}
\begin{enumerate}
  \item Initialize weights: $w_i = 1.0$ for each expert $i \in \{1, 2, ..., N\}$.
  
  \item At each round $t$:
  \begin{enumerate}
    \item Compute total weight: $W = \sum_{i=1}^{N} w_i$.
    \item Compute the probability for each expert:
    \[
      p_i = \frac{w_i}{W}, \quad \forall i \in \{1, ..., N\}.
    \]
    \item Sample one expert $i_t$ according to distribution $\{p_i\}$.
    \item Predict SQL using expert $i_t$: $\hat{s}_t \leftarrow \text{SQL candidate from } i_t$.
  \end{enumerate}

  \item After receiving supervision (e.g., gold SQL or execution result), update weights:
  \[
    w_i \leftarrow w_i \cdot (1 - \varepsilon)^{\ell_t(i)},
  \]
  where $\ell_t(i) = 1$ if expert $i$ makes a mistake at time $t$, and $0$ otherwise.


\item \textbf{Epsilon setting:} To minimize the expected mistake bound, set
  \[
    \varepsilon = \sqrt{\frac{\log N}{T}},
  \]

  \item \textbf{Mistake Bound:}
  \[
    \mathbb{E}[M_T] \leq (1 + \varepsilon) M^* + \frac{\log N}{\varepsilon} \leq M^* + 2\sqrt{T\log N}
  \]
      where $N$ is the number of experts, and $M^*$ is the number of mistakes made by the best expert during $T$ rounds.
\end{enumerate}
\end{algorithm}

\subsubsection{Execution-Guided Self-Refinement}
Inspired by ReFoRCE, we incorporate execution-aware correction strategies and design a simplified self-refinement loop for our SSEV pipeline. Unlike traditional methods that rely on explicit error categorization or external planning modules, our approach directly applies an execution-guided self-refinement loop when a generated SQL query fails due to syntax errors or produces empty results. The refinement process is performed for up to N iterations to reduce execution failures. At each iteration, the model is re-prompted with the previously generated SQL, its schema-aware context, and the corresponding execution error message. The process terminates early if the query executes successfully or stops upon reaching the maximum iteration limit. Importantly, SQL candidates that remain unsuccessful after refinement are still retained and considered in the final voting stage. In this way, the proposed self-refinement mechanism remains lightweight while effectively improving robustness.

\begin{figure}[htb!]
    \centering
    \includegraphics[width=\textwidth]{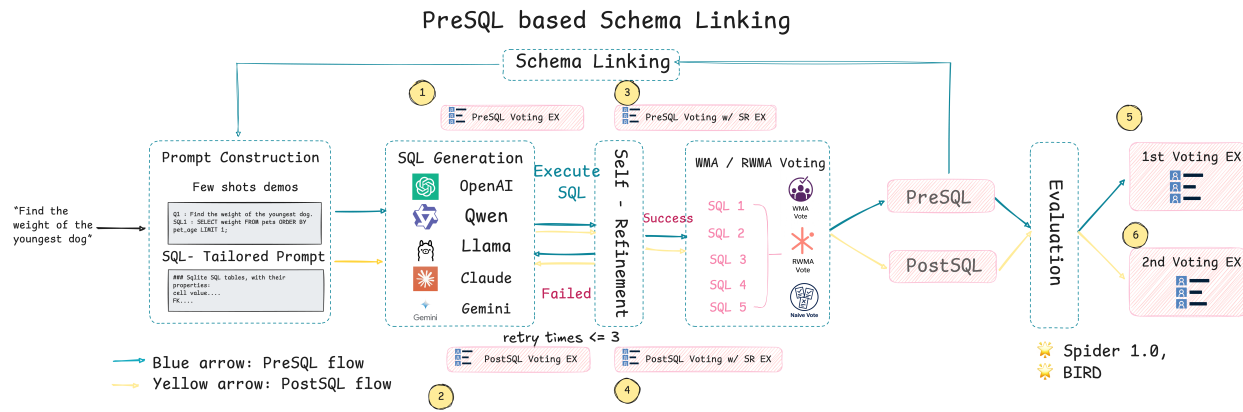}
    \caption{
        Architecture of the two-stage Text-to-SQL system: PreSQL (blue) and PostSQL (yellow) pipelines. Both stages involve prompting, schema linking, generation, voting (WMA, RWMA, or Naïve), and execution-guided Self-refinement.
    }
    \label{fig:single_agent_pipeline}
    
    \vspace{4pt}
    \small
\end{figure}

\subsection{SSEV pipeline}
\autoref{fig:single_agent_pipeline} illustrates the two-stage SSEV pipeline, consisting of the PreSQL and PostSQL stages, which progressively refine SQL generation through schema-aware prompting, execution-guided self-refinement, and adaptive voting. In the PreSQL stage, few-shot prompts guided by schema-aware similarity are provided to multiple LLMs to generate initial SQL candidates, followed by schema linking and execution-guided self-refinement to mitigate errors. In the PostSQL stage, a pruned, schema-informed prompt is used for a second round of SQL generation, where candidates are further refined and evaluated. 

In \autoref{fig:single_agent_pipeline}, blue lines and arrows denote the PreSQL workflow, while yellow lines and arrows represent the PostSQL workflow.  In both stages, a voting mechanism aggregates predictions from multiple LLMs, while execution-guided self-refinement iteratively improves SQL quality based on execution feedback. To assess voting strategies, we compare the proposed WMA and RWMA methods with a naïve approach that selects the most frequently generated SQL, treating all experts equally without considering past accuracy or domain expertise.

\begin{figure}[htb]
    \center
    \includegraphics[width=1.0\textwidth]{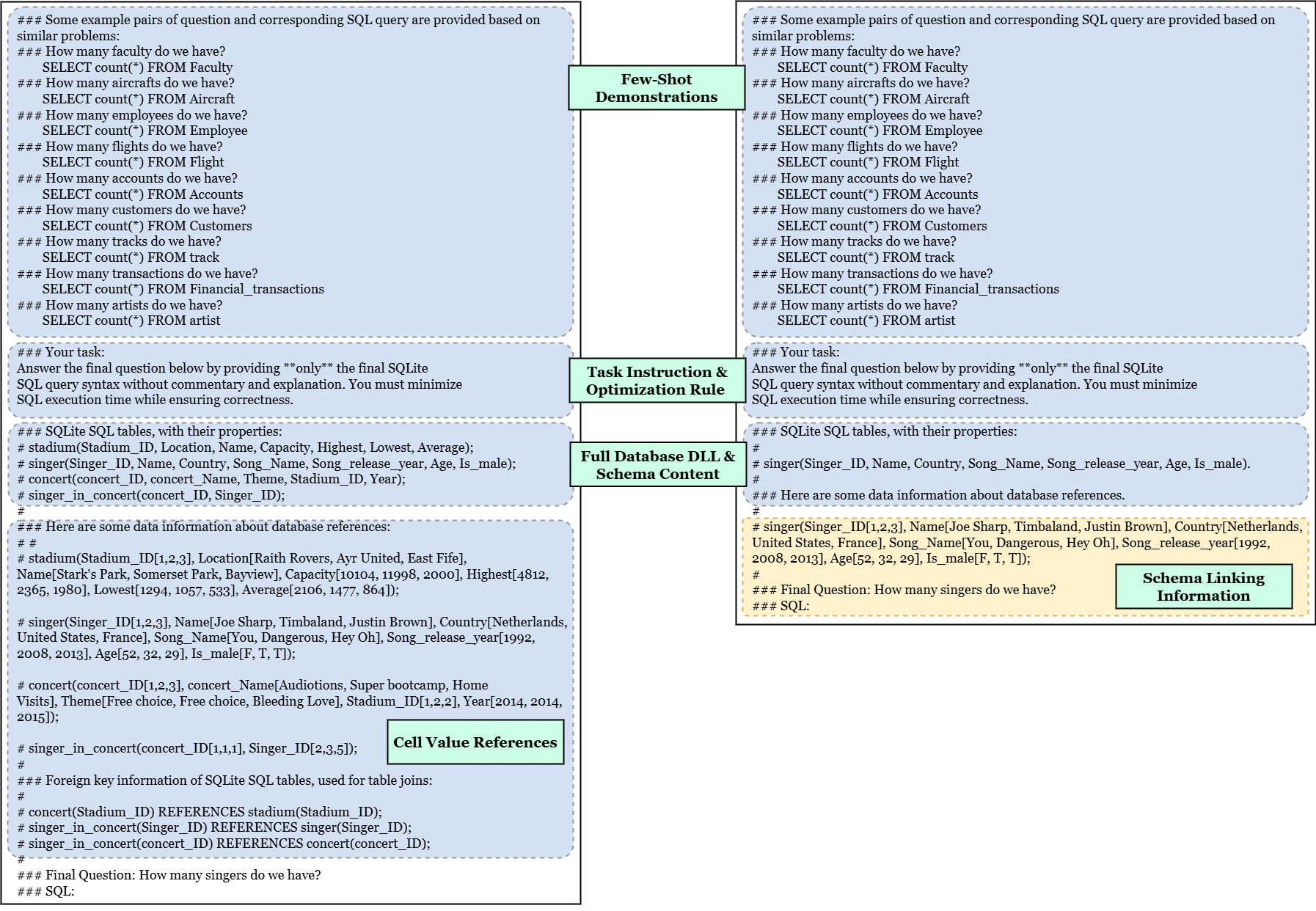}
    \caption{
        Comparison of PreSQL and PostSQL prompt structure. The left side illustrates the PreSQL prompt, which includes complete
schema DDL, full table cell references, and few-shot examples to guide initial SQL generation. The right side shows the PostSQL prompt, where the schema has been pruned to only include entities linked in the PreSQL output, enabling more focused and efficient SQL generation.
    }
    \label{fig:presql_postsql_combined}
\end{figure}

\begin{figure}[htb]
    \center
    \includegraphics[width=0.5\textwidth]{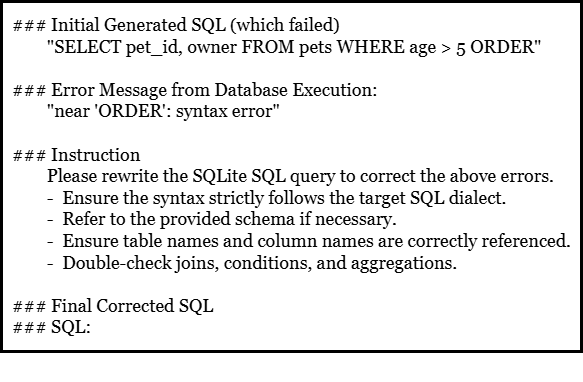}
    \caption{Refinement prompt example
    }
    \label{fig:Refinement Prompt}
\end{figure}

The key components of the SSEV pipeline are summarized as follows:

\begin{itemize}
    \item \textbf{Prompt Construction} \\
    We begin by embedding the user’s natural language question into a SQL-tailored few-shot prompt (\autoref{fig:presql_postsql_combined}), which includes demonstration pairs and lightweight schema hints.
    \item \textbf{PreSQL Generation}  \\
    Multiple LLMs (e.g. GPT-4, LLaMa, Gemini) are triggered to generate preliminary SQL statements (“PreSQL”).
    \item \textbf{Schema Linking} 
    Each PreSQL is parsed to extract referenced tables and columns, and unrelated schema elements are pruned.
    \item \textbf{Prompt Refinement}  \\
    The linked schema is combined into the prompt template, producing a concise PostSQL prompt. (see \autoref{fig:presql_postsql_combined})
    \item \textbf{PostSQL Generation}  \\
    The same ensemble of LLM experts generates refined SQL candidates (“PostSQL”) using the refined prompt.
    \item \textbf{Self-Refinement}  \\
    Self-Refinement: SQL queries that fail execution due to syntax errors or empty results undergo up to N refinement iterations using their prior schema-aware context and the error message. Failed queries are still included in final voting.
    
        \begin{itemize}
            \item \textbf{Trigger Conditions:} A query enters refinement if execution returns a syntax error or empty result.
            \item \textbf{Iterative Rewrite Strategy:} Each failed SQL is re-prompted with its previous prompt, schema context, and last database error; the LLM generates a revised query, repeated up to M times. The rewritten prompt is illustrated in the \autoref{fig:Refinement Prompt}.
            \item \textbf{Termination Criteria:} Refinement stops on successful execution or after M attempts; remaining failures are included in voting.
        \end{itemize}

    \item \textbf{WMA/RWMA/Naïve Voting} 
    Finally, all valid PreSQL and PostSQL outputs are weighted via an adaptive voting strategy to select the best query.
    
\end{itemize}

\subsection{ReCAPAgent-SQL}
Although ReCAPAgent-SQL builds upon insights from our SSEV pipeline, it represents a substantially more comprehensive and structurally distinct framework. ReCAPAgent-SQL is not only a multi-agent system, but also systematically incorporates the core ideas of PreSQL generation, schema linking, PostSQL refinement, execution-guided self-refinement, and expert voting by encapsulating them within specialized agents. In addition, the framework integrates explicit validation and information retrieval components to support more robust reasoning and reliable execution.

\begin{figure}[htb]
    \flushleft
    \includegraphics[width=1.0\textwidth]{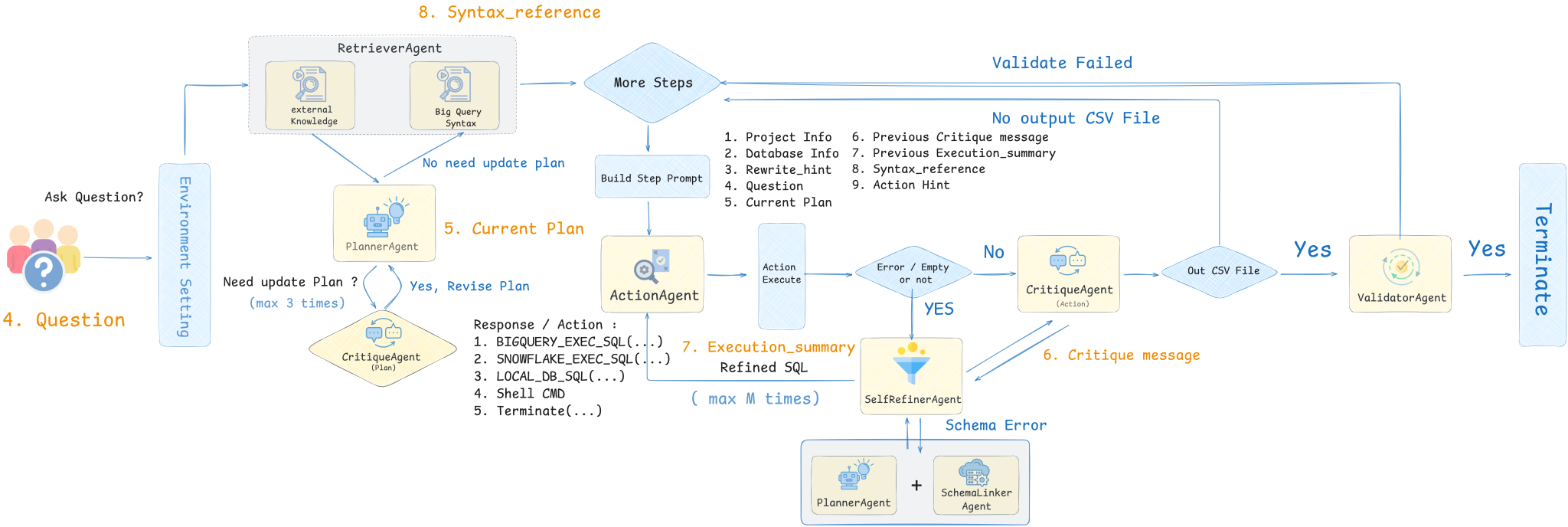}
    \caption{
        Architecture of ReCAPAgent-SQL with core agent modules.
    }
    \label{fig:recap_architecture}
    
    \vspace{4pt}
    \small
\end{figure}

As illustrated in \autoref{fig:recap_architecture}, ReCAPAgent-SQL comprises several cooperating agents, including the PlannerAgent, RetrieverAgent, CritiqueAgent, SchemaLinkerAgent, ActionAgent, SelfRefinerAgent, and ValidatorAgent, which together enable iterative SQL refinement. The workflow begins with a user-provided natural language query and environment settings, followed by the construction of a multi-step reasoning plan by the PlannerAgent. This plan is iteratively critiqued and refined by the CritiqueAgent for up to three iterations. Based on the constructed prompts, the ActionAgent predicts and executes the corresponding actions. Execution failures or empty results trigger the SelfRefinerAgent to regenerate refined SQL candidates, while schema-related errors invoke collaboration between the SchemaLinkerAgent and PlannerAgent for dynamic schema resolution. Finally, the ValidatorAgent verifies the execution results, and the process terminates upon successful validation. Collectively, these agents support plan decomposition, iterative refinement, dynamic schema understanding, and execution verification, enabling ReCAPAgent-SQL to effectively address complex, multi-step Text-to-SQL tasks.

The framework loop through the plan–execute–critique–refine cycles until a valid and complete result is obtained or a predefined maximum step limit is reached. Table~\ref{tab:agent_responsibilities} summarizes the activation conditions and responsibilities of each specialized agent in the ReCAPAgent-SQL framework, highlighting their interactions within the plan–execute–refine–validate cycle.

\begin{table}[htb!]
\centering
\small
\caption{Responsibilities and Trigger Conditions of Each Agent in ReCAPAgent-SQL}
\label{tab:agent_responsibilities}
\begin{tabular}{p{3.3cm} p{6.2cm} p{4.5cm}}
\toprule
\textbf{Agent} & \textbf{Responsibility} & \textbf{Trigger Condition} \\
\midrule
\textbf{External Knowledge RetrieverAgent} & Provides external knowledge to enrich the planning prompt & Before plan generation \\

\textbf{PlannerAgent} & Generates a step-by-step reasoning plan & After environment setup and external knowledge retrieval \\

\textbf{PlanCritiqueAgent} & Evaluates and revises the initial plan (up to three times) & Immediately after plan generation \\

\textbf{Syntax RetrieverAgent} & Supplies SQL syntax reference to the prompt if needed & After planning, when syntax is requested \\

\textbf{ActionAgent} & Predicts the next executable action (SQL, Bash, Terminate) & At each step in the execution loop \\

\textbf{SQLCritiqueAgent} & Reviews SQL and provides feedback or revision suggestions & After each SQL generation step \\

\textbf{SelfRefinerAgent} & Refines failed SQL outputs iteratively & Upon execution error or empty result \\

\textbf{SchemaLinkerAgent} & Injects schema-related information into the SQL prompt & When schema-related errors occur (e.g., column/table not found) \\

\textbf{ValidatorAgent} & Verifies final SQL output against intent and format & After successful SQL execution \\
\bottomrule
\end{tabular}
\end{table}

\subsubsection{PlannerAgent: Task Decomposition \& Prompt Format}
The PlannerAgent generates a step-by-step reasoning plan to guide SQL query prediction. Instead of producing SQL directly, it first constructs a reasoning plan, guiding subsequent modules generate more accurate queries. This approach enables the LLM to better understand diverse databases and SQL dialects. With a reasoning plan, downstream agents can effectively apply critique and self-refinement mechanisms.

\begin{figure}[htb]
    \center
    \includegraphics[width=0.9\textwidth]{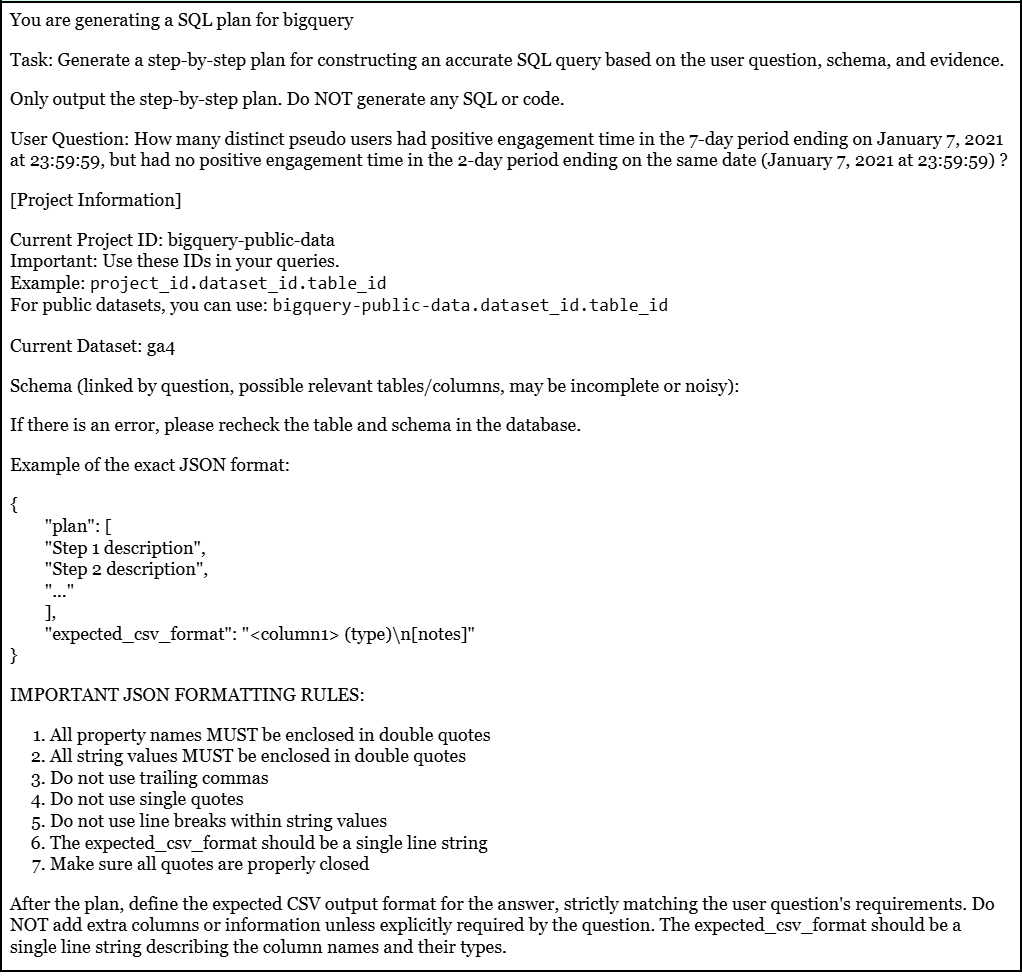}
    \caption{Generate reference plan prompt
    }
    \label{fig:Generate_Reference_Plan_Prompt}
\end{figure}

\begin{figure}[htb]
    \center
    \includegraphics[width=0.9\textwidth]{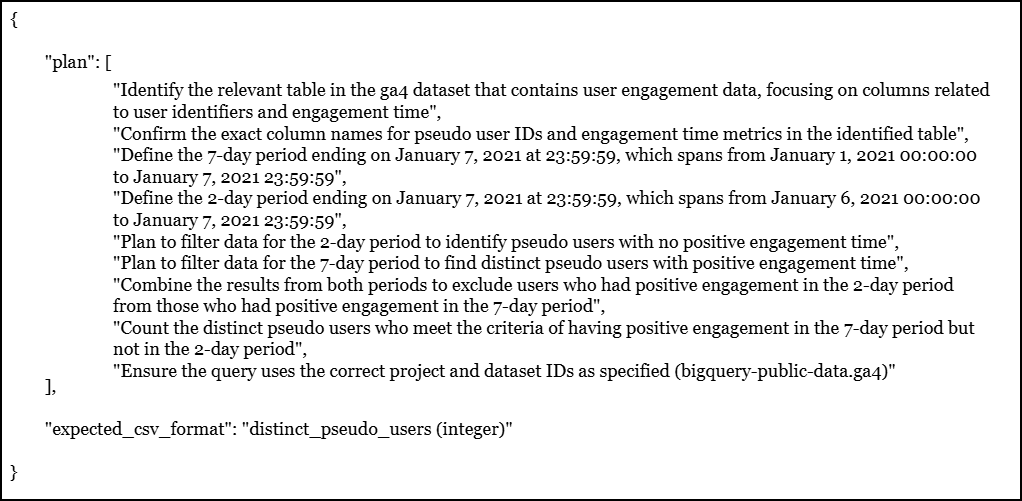}
    \caption{Reference plan example
    }
    \label{fig:Reference_Plan_Example}
\end{figure}

\begin{itemize}

    \item \textbf{Prompt Format}  \\
    The PlannerAgent constructs a carefully designed prompt with the following key elements (see \autoref{fig:Generate_Reference_Plan_Prompt} and \autoref{fig:Reference_Plan_Example}):
    \begin{itemize}
        \item \textbf{Project Information:} For dialects such as BigQuery or Snowflake, the prompt includes project and dataset context to avoid errors in table reference or permission denied.
        \item \textbf{User Question:} The natural language question to be answered.
        \item \textbf{Task Instruction:} The agent explicitly instructs the LLM to produce \textbf{only} a step-by-step plan in strict JSON format.
        \item \textbf{Expected CSV Format:} The agent also asks the LLM to define the expected output format (column names and types).
    \end{itemize}

     \item \textbf{Response Parsing}  \\
     The LLM's response is parsed into a \texttt{plan} (a list of steps) and an \texttt{expected\_csv\_format}. To ensure robustness, the agent is enforced to output a format that the subsequent agents can follow.

\end{itemize}

\subsubsection{RetrieverAgent Agent: External Knowledge and BigQuery Syntax Retrieval}
Inspired by DB-GPT \cite{Xue2023DBGPT}, we implement a lightweight RAG mechanism to retrieve relevant information for SQL generation. ReCAPAgent-SQL uses two RAG components:

\begin{itemize}
    \item\textbf{External Knowledge RetrieverAgent}  \\
    Retrieves document snippets relevant to the user query using dense vector search (SentenceTransformer embeddings + FAISS). It indexes an external document corpus and fetches top-k relevant snippets to provide domain knowledge context during SQL generation

    \item\textbf{Syntax RetrieverAgent for BigQuery}  \\
    SQL dialect–specific syntax (e.g., BigQuery) often includes specialized function parameters or advanced usage patterns that may not be fully represented in an LLM’s training corpus. The Syntax RetrieverAgent module addresses this limitation by retrieving relevant syntax documentation from a dedicated syntax repository. Whether this module is invoked is determined by the Syntax RetrieverAgent itself: it prompts the LLM to assess whether additional dialect-specific clarification is required (see \autoref{fig: Syntax RetrieverAgent Prompt}).

    \begin{figure}[htb]
      \centering
      \includegraphics[width=0.5\textwidth]{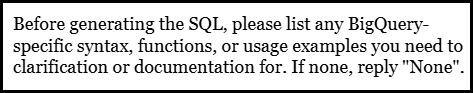}
      \caption{Syntax RetrieverAgent Prompt Example.}
      \label{fig: Syntax RetrieverAgent Prompt}
    \end{figure}

\end{itemize}

\subsubsection{Critique Agent: Plan/SQL Evaluation Strategy}
The Critique Agent provides automated evaluation and refinement for SQL generation, improving robustness via iterative critique.

\begin{itemize}

    \item\textbf{Plan Critique}   \\
    Given a step-by-step reasoning plan produced by the PlannerAgent, the Critique Agent conducts a structured evaluation of the plan by prompting an LLM to assess (1) whether the plan includes all necessary steps for the SQL agent to answer the question, and (2) whether the reasoning is clear and coherent. The LLM returns its assessment in a strict JSON format, indicating whether plan refinement is required (\texttt{update\_plan}). If refinement is needed, the system triggers additional iterations of the PlannerAgent to revise the plan accordingly.

    \item\textbf{SQL Critique}   \\
    After SQL generation, the Critique Agent evaluates the query for correctness, compliance with the target SQL dialect (e.g., BigQuery, Snowflake, SQLite), and consistency with the original reasoning plan. Guided by a structured prompt, the LLM identifies and explains potential errors or inconsistencies and provides a revised SQL query that addresses these issues. The response follows a two-part format:
        \begin{enumerate}
            \item \texttt{[Reasoning]} — An explanation of detected flaws.
            \item \texttt{[SQL]} — The corrected query, wrapped in an executable \texttt{Action} block.
        \end{enumerate}
\end{itemize}

By systematically critiquing both the reasoning plan and the SQL output, the Critique Agent enhances the overall accuracy, reliability, and interpretability of our ReCAPAgent-SQL pipeline.

\subsubsection{SchemaLinkerAgent: Dynamic Column Filtering}
We introduce a SchemaLinkerAgent that dynamically derives a concise, context-aware schema representation from raw database documentation. By narrowing the schema to only relevant tables and columns, this module reduces noise and mitigates errors caused by irrelevant schema elements.

\begin{itemize}

    \item\textbf{Schema Filtering}  \\
    When execution errors such as \textit{TableNotFound} or \textit{ColumnNotFound} occur, schema filtering is triggered. The PlannerAgent dynamically prunes the database schema to retain only relevant components. If a set of \textit{linked tables} or \textit{linked columns} is available from prior schema linking steps, only these elements are included in the prompt, ensuring that the LLM focuses on pertinent schema information.
        
    \item\textbf{Agent Process}   \\
    The \texttt{SchemaLinkerAgent} operates in an iterative loop, using a Docker environment to execute CLI commands (e.g., \texttt{ls}, \texttt{cat}, \texttt{head}, \texttt{grep}) for step-by-step exploration. It examines schema resources such as "DDL.csv", "JSON files", "sample CSV data", and textual documentation to extract table structures, column names, data types, and sample values. To prevent infinite exploration, the process is capped at a fixed number of steps (default: 30).
    
    \item\textbf{Schema Formatting} \\  
    After collecting sufficient schema-related information, the agent issues a \texttt{Terminate} action and outputs a compact, structured schema representation:
        \begin{quote}
            \texttt{table1(col1:TYPE[val1,val2], col2:TYPE[val3,val4]); table2(...)}
        \end{quote}
    This format is tailored to optimize downstream PlannerAgent reasoning and subsequent SQL generation.

\end{itemize}

\subsubsection{Action Agent: Action Prediction}
\begin{figure}[htb]
      \centering
      \includegraphics[width=0.6\textwidth]{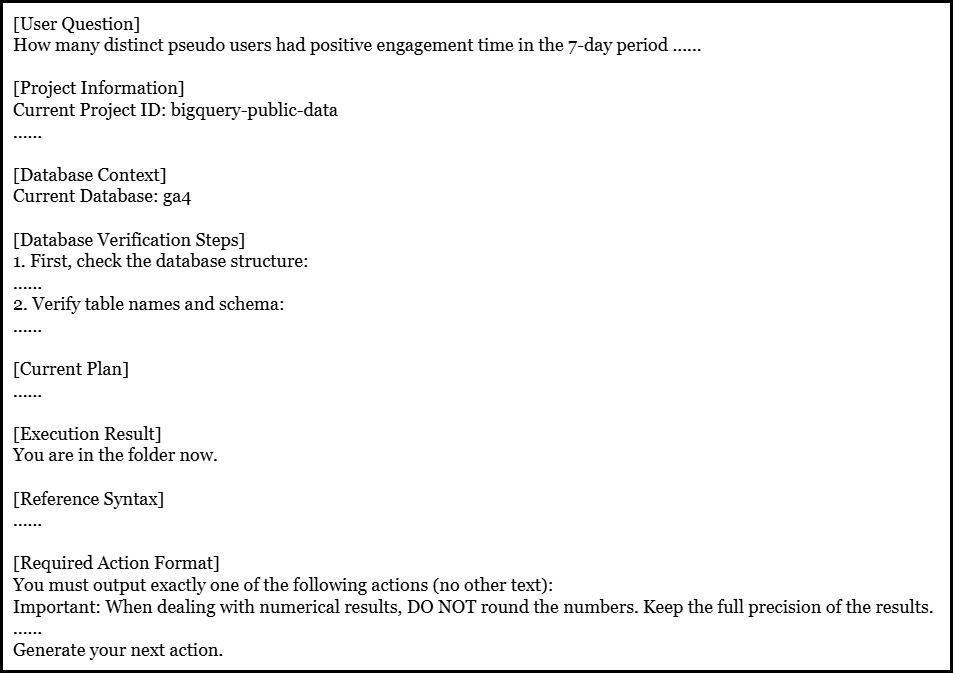}
      \caption{Predict Action Prompt: Given the user question and environment, the LLM is instructed to generate an accurate SQL execution plan, referencing schema and timestamps.}
      \label{fig:predict_action_prompt}
    \end{figure}

The Action Agent dynamically predicts the next executable action based on the current state of the ReCAPAgent-SQL pipeline. It operates in a step-wise manner, leveraging the structured reasoning plan produced by the PlannerAgent and feedback from previous iterations (see Figure~\ref{fig:predict_action_prompt}).

\begin{itemize}

    \item\textbf{Prompt-driven Action Prediction}   \\
    At each iteration, the Action Agent constructs a comprehensive prompt that integrates several key elements: (1) the original user question, (2) the current database and project context, (3) the current step from the PlannerAgent's reasoning plan, (4) feedback from the most recent Critique Agent evaluation of the prior action (if available), (5) execution results or error messages from the previous step, (6) retrieved dialect-specific syntax to guide SQL generation, and (7) strict output constraints to ensure valid executable actions.
        
    \item\textbf{Iterative Prediction Loop}  \\
    The Action Agent interacts with the LLM through an iterative loop using the \texttt{predict(obs)} method. At each step, (1) the latest \texttt{observation} is appended to the message history, (2) the LLM predicts a structured response consisting of a \texttt{Thought} and an \texttt{Action}, (3) the system parses the response to extract the action, validates it, and updates its internal historical log, and (4) when necessary, error-handling mechanisms invoke the SelfRefinerAgent to recover from malformed outputs or execution errors.
        
    \item\textbf{Controllability and Safety}   \\
    The Action Agent enforces strict constraints on allowable output actions (e.g., \texttt{BIGQUERY\_EXEC\_SQL}, \texttt{Terminate}) to ensure controllable execution. Repetitive actions and invalid outputs are detected so that they won't be executed.

    \item\textbf{Role in ReCAPAgent-SQL Agent Loop}   \\
    The Action Agent is a core component of the ReCAPAgent-SQL Agent’s execution loop. After the PlannerAgent generates a reasoning plan, each plan step is processed by the Action Agent to produce executable SQL fragments or explorable actions. Moreover, after each execution, the Critique Agent evaluates the execution result, and the Action Agent incorporates this feedback into subsequent predictions. This enables a fully adaptive, critique-driven refinement loop for robust SQL generation.

\end{itemize}

\subsubsection{SelfRefinerAgent: Feedback Loop \& Error Correction}
The SelfRefinerAgent is considered the core element of our system’s iterative feedback loop for robust SQL generation because it enables the pipeline to dynamically analyze execution failures and employ LLM-guided refinement to modify and improve SQL queries across multiple iterations.

    \begin{figure}[htb!]
      \centering
      \includegraphics[width=0.6\textwidth]{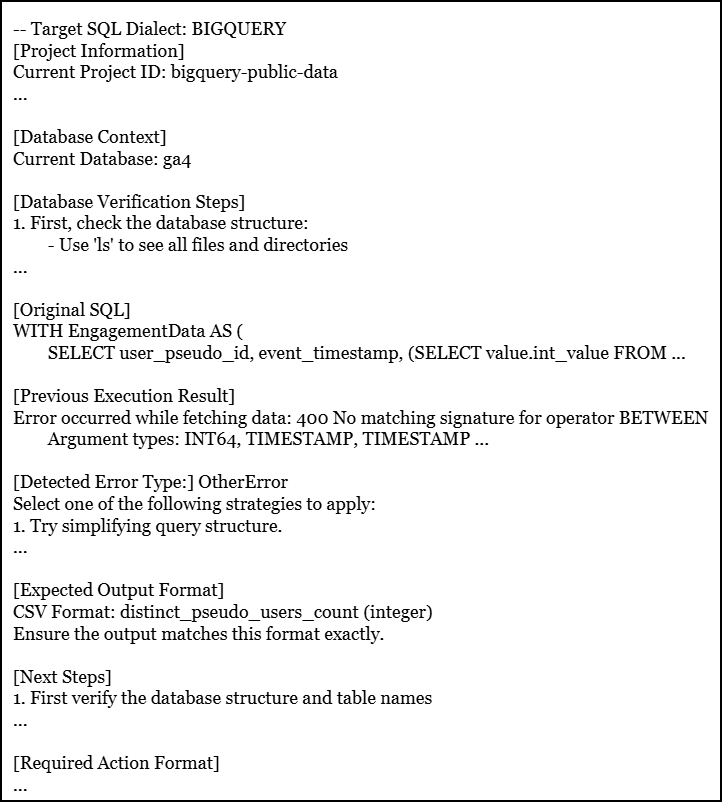}
      \caption{Self-Refinement Prompt: Guides the LLM in analyzing the SQL failure reason and refining the query using a list of suggested actions, error messages, and schema context.}
      \label{fig:self_refine_prompt_fix}
    \end{figure}

\begin{itemize}

    \item\textbf{Loop Structure}   \\
    At each iteration, the SelfRefinerAgent executes the following steps:
    \begin{itemize}
        \item \textbf{Error Detection}: The system classifies execution outcomes into distinct categories, such as SQL errors, empty results, or unknown issues.
        \item \textbf{Error Type Analysis}: Using the \texttt{construct\_error\_prompt} module, the agent maps detected errors to a predefined taxonomy (e.g., \texttt{SyntaxError}, \texttt{TableNotFound}, \texttt{TypeMismatch}, \texttt{NoResult}) and incorporates detailed error-specific hints.
        \item \textbf{Dynamic Prompt Construction}: A comprehensive refinement prompt is constructed by integrating several key elements: (1) the original SQL query, (2) execution observations and error messages, (3) prior critique feedback, (4) error type hints along with suggested correction strategies, (5) any available schema linking information, (6) the expected output format, and (7) relevant dialect-specific constraints and required action format. An example of such a prompt can be seen in \autoref{fig:self_refine_prompt_fix}.
        \item \textbf{LLM-driven Refinement}: The agent invokes an LLM to predict a refined SQL query or terminate the loop if correction is deemed successful.
        \item \textbf{Execution \& Critique}: The refined query is executed, and the Critique Agent evaluates the result. If errors persist, the loop continues until reaching the maximum refinement limit.
    \end{itemize}
        
    \item\textbf{Termination Criteria}  \\
    The loop terminates under any of the following conditions: (1) the refined SQL query executes successfully without errors, (2) the result is non-empty and matches the expected output format, or (3) the maximum number of self-refinement iterations has been reached.

\end{itemize}

\begin{figure}[H]
  \centering
  \includegraphics[width=0.8\textwidth]{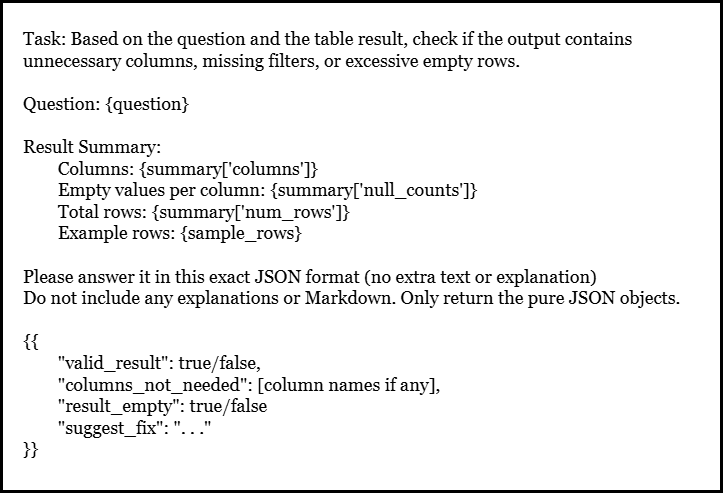}
  \caption[Validate Result Prompt of ReCAPAgent-SQL]{Validate Result Prompt: Checks whether the query result is correct, empty, or contains redundant columns. Returns a structured JSON verdict.}
  \label{fig:validate_prompt}
\end{figure}

\subsubsection{ValidatorAgent: Result Validation \& Termination Criteria}
The ValidatorAgent acts as the final quality assurance checkpoint in the ReCAPAgent-SQL pipeline. Its role is to ensure that the generated SQL query produces results aligned with the original user question before terminating the ReCAPAgent-SQL loop. This process helps prevent errors such as over-generation, hallucinated columns, incomplete filters, or empty outputs.

\begin{itemize}

    \item\textbf{Validation Process} \\  
    The ValidatorAgent constructs a structured prompt presenting the original user question along with a summary of the result table, including column names, null counts per column, number of rows, and sample rows (i.e., first three rows) (see Figure~\ref{fig:validate_prompt}). The LLM is then prompted to evaluate the result with the following objectives: (1) identify any unnecessary columns, (2) detect missing filters or incomplete logic, (3) determine whether the result is empty or uninformative, and (4) provide suggested fixes or improvement strategies if any issues are found.

    \item\textbf{Response Format} \\  
    To ensure reliable parsing and downstream consistency, the LLM outputs its evaluation in a strict JSON format:
    
    \begin{quote}
    \small
    \texttt{\{"valid\_result": true/false, "columns\_not\_needed": [...], "result\_empty": true/false, "suggest\_fix": "..." \}}
    \end{quote}
        
    \item\textbf{Termination Criteria} \\  
    The output of the ValidatorAgent governs whether the refinement loop terminates successfully. The loop proceeds as follows: (1) if the result passes validation and is non-empty, the system triggers a \texttt{Terminate} action; (2) if the result is invalid, empty, or contains unnecessary columns, the system re-enters the ReCAPAgent-SQL loop for further prediction.
\end{itemize}

\section{Evaluation}
\label{ch:experiments}
We evaluate the performance of the SSEV pipeline using the Spider 1.0 and BIRD benchmark datasets, and we further investigate the ReCAPAgent-SQL system using the Spider 2.0-lite dataset. These datasets encompass a wide variety of domains, schema structures, and SQL complexities. Execution accuracy (EX) is employed as the primary evaluation metric.

Spider 1.0 provides a foundational cross-domain benchmark but is limited by small tables and lack of realistic database values. BIRD introduces real-world challenges, including noisy data and domain-specific knowledge, containing 12,751 examples across 95 large databases in 37 domains. Unlike the prior benchmarks, Spider 2.0 targets enterprise workflows, long-context reasoning, dialect diversity, and SQL generation in dynamic environments (e.g., BigQuery, Snowflake), featuring over 800 columns and queries exceeding 100 tokens. These benchmarks collectively highlight the shift from synthetic, structured datasets to realistic, dialect-sensitive environments, motivating systems that support multi-turn reasoning and execution-aware SQL generation.

All experiments were conducted on a local machine with an NVIDIA RTX 4090 GPU (24GB VRAM) and 64GB RAM, implemented in Python 3.10. We evaluated the framework with eight LLMs using unified prompt templates and official APIs: Llama-3.3-70B \cite{grattafiori2024llama3}, Qwen2.5-72B \cite{qwen2024hf}, Qwen2.5-coder-32B \cite{hui2024qwen2_5coder} (Together AI), GPT-4o \cite{openai2024gpt4o}, GPT-4.1-2025 \cite{openai2025gpt41}, o3-mini (OpenAI) \cite{openai2025o3mini}, Gemini-2.5-Pro Experimental (Google AI Studio) \cite{deepmind2025gemini2.5}, and Grok-3-beta (xAI) \cite{xai2025grok3}.

\subsection{SSEV Pipeline} 
To investigate the influence of various components in the SSEV pipeline, six experimental stages were designed to evaluate their impact.
\begin{itemize}
    \item \textbf{Stage 1 (Base)}: This stage examines the impact of applying ensemble voting (WMA, RWMA, or Naïve) alone, without schema linking or execution-guided self-refinement, serving as the baseline.
    \item \textbf{Stage 2 (Base + SL)}: This stage extends the baseline by incorporating schema linking. The voting (WMA, RWMA, or Naïve) is performed on schema-linking outputs.
    \item \textbf{Stage 3 (Base + SR)}: We add execution-guided self-refinement, applying voting (WMA, RWMA, or Naïve) after the refinement process.
    \item \textbf{Stage 4 (Base + SR + SL)}: We combines both execution-guided self-refinement and schema linking, with voting (WMA, RWMA, or Naïve) applied after both modules.
    \item \textbf{Stage 5 (Base + SR + Vote)}: We apply voting (WMA, RWMA, or Naïve) after execution-guided self-refinement, using the voting results for schema-linking.
    \item \textbf{Stage 6 (Base + SR + Vote + SL + SR + Vote)}: We implement alternative voting strategies (WMA, RWMA, or Naïve) after schema linking and execution-guided self-refinement, following the same procedure as Stage 5.
    
\end{itemize}

\subsubsection{Baseline}
Based on the six stages described above, we establish baseline results by evaluating each LLM on the Spider 1.0 and BIRD datasets in Stage 1. Building on these baselines, ensemble voting strategies (i.e., WMA, RWMA, and Naïve) are further evaluated in combination with execution-guided self-refinement across the remaining four progressive stages. All models are evaluated under identical 9-shot in-context learning settings. \autoref{tab:baseline_across_llms} shows the performance on each LLM and Gemini-2.5-Pro Experimental perform better than other models across three different datasets. 

\begin{table}[htb!]
    \centering
    \caption{Execution accuracy of baseline methods across multiple LLMs on Spider 1.0 and BIRD datasets}
    \begin{tabular}{llcccccc}
        \toprule
        \textbf{Dataset} & \textbf{Stage} 
        & \textbf{Llama-3.3-70B} 
        & \textbf{Qwen2.5-72B} 
        & \textbf{*Qwen2.5-32B} 
        & \textbf{GPT-4o} 
        & \textbf{o3-mini} 
        & \textbf{*Gemini-2.5} \\
        \midrule
        Spider 1.0 Dev & 1        & 77.3\% & 79.0\% & 82.3\% & 79.9\% & 81.7\% & 84.9\% \\
        \midrule
        Spider 1.0 Test  & 1  & 78.6\% & 80.9\% & 80.9\% & 84.0\% & 83.2\% & 85.7\% \\
        \midrule
        BIRD   & 1    & 61.41\% & 57.56\% & 60.37\% & 64.02\% & 61.15\% & 65.97\% \\
        \midrule
        \multicolumn{8}{p{15cm}}{\footnotesize
        \textit{Note:} \par 
        Qwen2.5-32B: Qwen2.5-coder-32B-instruct-fp16 \par 
        Gemini-2.5:  Gemini-2.5-Pro Experimental
        } \\
        \bottomrule
    \end{tabular}
    \label{tab:baseline_across_llms}
\end{table}

\subsubsection{Cross Stage Investigation on Spider 1.0 Dev}
\autoref{tab:spider1_dev_results} reports execution accuracy on the Spider 1.0 development set across the six incremental stages. Overall, we observe consistent performance improvements as schema linking, execution-guided self-refinement, and ensemble voting strategies are progressively incorporated into the pipeline.

\begin{table}[htb!]
\centering
\caption{Execution accuracy (Spider 1.0 Dev) - voting applied at each stage}
\begin{tabular}{l|ccc|cccccc}
\toprule
Stage  & WMA & RWMA & Naïve & *Llama & *Qwen2.5-1 & *Qwen2.5-2 & GPT-4o & o3-mini & *Gemini-2.5 \\
\midrule
1.   & 0.846 & 0.814 & 0.796 & 0.773 & 0.790 & 0.823 & 0.799 & 0.817 & 0.849 \\
2.   & 0.844 & 0.816 & 0.803 & 0.776 & 0.802 & 0.826 & 0.810 & 0.817 & 0.848 \\
3.   & 0.839 & 0.828 & 0.809 & 0.784 & 0.797 & 0.829 & 0.799 & 0.817 & 0.841 \\
4.   & 0.844 & 0.824 & 0.795 & 0.770 & 0.799 & 0.819 & 0.813 & 0.819 & 0.848 \\
5.   & 0.839 & 0.828 & 0.809 & 0.784 & 0.797 & 0.829 & 0.799 & 0.817 & 0.841 \\
6.1  & \textbf{0.855} & -- & -- & 0.789 & 0.809 & 0.831 & 0.806 & 0.818 & \textbf{0.856} \\
6.2  &  -- & 0.829 & -- & 0.787 & 0.815 & 0.822 & 0.806 & 0.826 & 0.844 \\
6.3  & -- & -- & 0.797 & 0.781 & 0.811 & 0.821 & 0.797 & 0.818 & 0.839 \\
\midrule
        \multicolumn{10}{p{15cm}}{\footnotesize
        \textit{Note:} \par 
        Llama: Llama3.3
        Qwen2.5-1:Qwen2.5-72B \par
        Qwen2.5-2: Qwen2.5-coder-32B-instruct-fp16 \par 
        Gemini-2.5:  Gemini-2.5-Pro Experimental \par
        Stage 6.1: with WMA Vote \par 
        Stage 6.2: with RWMA Vote \par
        Stage 6.3: with Naïve Vote
        } \\

\bottomrule
\end{tabular}

\label{tab:spider1_dev_results}
\end{table}
At the baseline level (see \autoref{tab:spider1_dev_results}, ensemble voting alone (Stage 1) already outperforms individual LLM. Introducing schema linking in Stage 2 yields modest performance gains, particularly for weaker models such as LLaMA-3.3 and Qwen2.5-72B. When execution-guided self-refinement is applied (Stages 3–4), the marginal benefit of schema linking becomes less impactful.

Among all configurations, Stage 6.1, which integrates WMA voting, execution-guided self-refinement, and schema linking, achieves the highest execution accuracy of 85.5\%, outperforming all other stages. This result highlights the effectiveness of our proposed ensemble-refinement framework.

\begin{figure}[htb!]
\centering
\includegraphics[width=\textwidth]{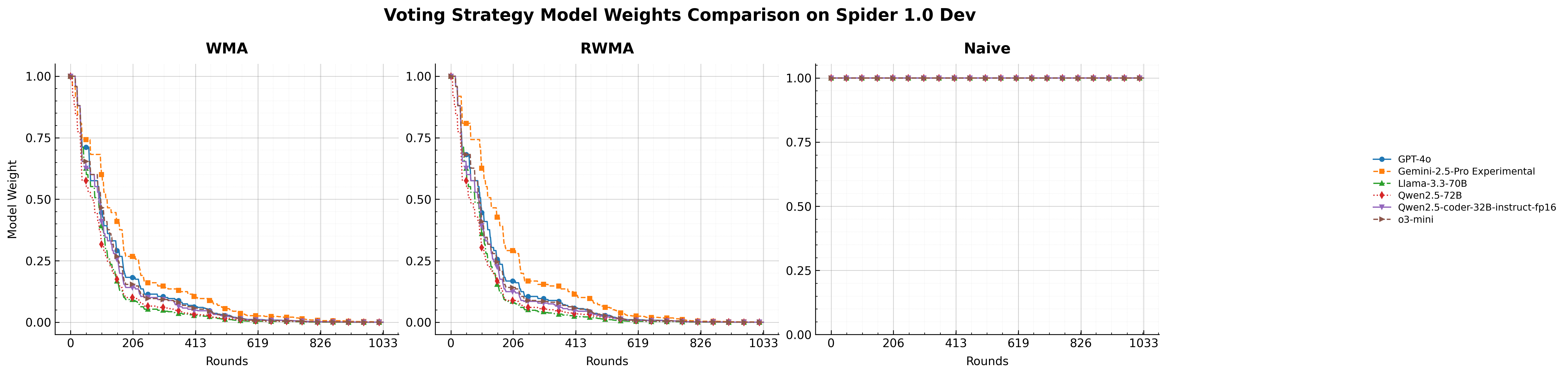}
\caption{Voting strategy model weights (Stage 6) on Spider 1.0 Dev. WMA shows rapid convergence to stronger models.}
\label{fig:spider1_dev_weights_best}
\end{figure}

\autoref{fig:spider1_dev_weights_best} illustrates the voting dynamics at Stage 6.1. WMA demonstrates rapid convergence, progressively assigning higher weights to stronger models within the first 200 rounds. Notably, Gemini-2.5-Pro Experimental consistently dominates the ensemble weights from Stage 1 through Stage 6.

\begin{figure}[htb!]
\centering
\includegraphics[width=\textwidth]{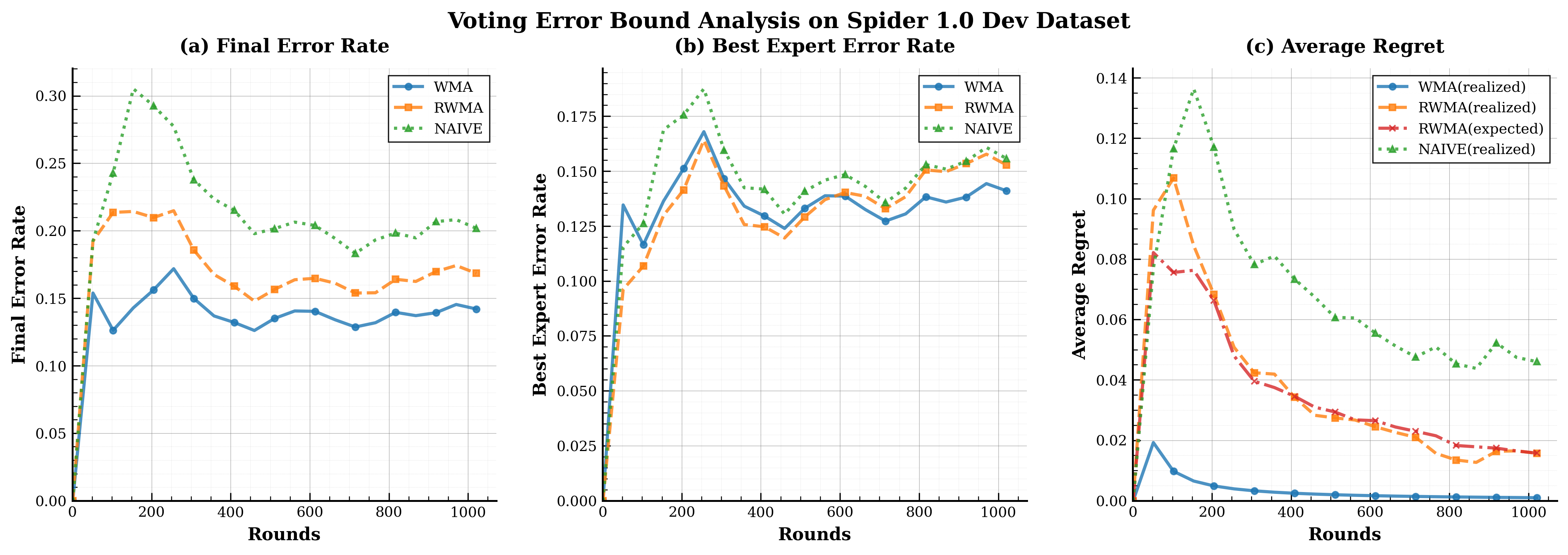}
\caption{Voting error bound analysis (Stage 6) on Spider 1.0 Dev. WMA achieves the lowest final error rate and regret.}
\label{fig:spider1_dev_error_best}
\end{figure}

As shown in \autoref{fig:spider1_dev_error_best}, WMA achieves significantly lower final error rates and average regret compared to RWMA and Naïve voting. In particular, the average regret of WMA closely approaches the error rate of the best-performing expert, indicating that WMA effectively adapts its weighting strategy toward the most accurate model over time. These results demonstrate that WMA not only stabilizes ensemble decision-making but also enables near expert-level performance in dynamic ensemble settings.

\subsubsection{Cross Stage Investigation on Spider 1.0 Test}
\autoref{tab:spider1_test_results} presents execution accuracy on the Spider 1.0 test set across the six incremental stages. The observed trends are consistent with those on the dev dataset, confirming the strong generalization capability of our proposed refinement voting components. As well, introducing schema linking in Stage 2 yields modest improvements for lower-performing models, while the addition of execution-guided self-refinement in Stage 3 further strengthens the robustness of the ensemble voting component.

\begin{figure}[htb!]
\centering
\includegraphics[width=\textwidth]{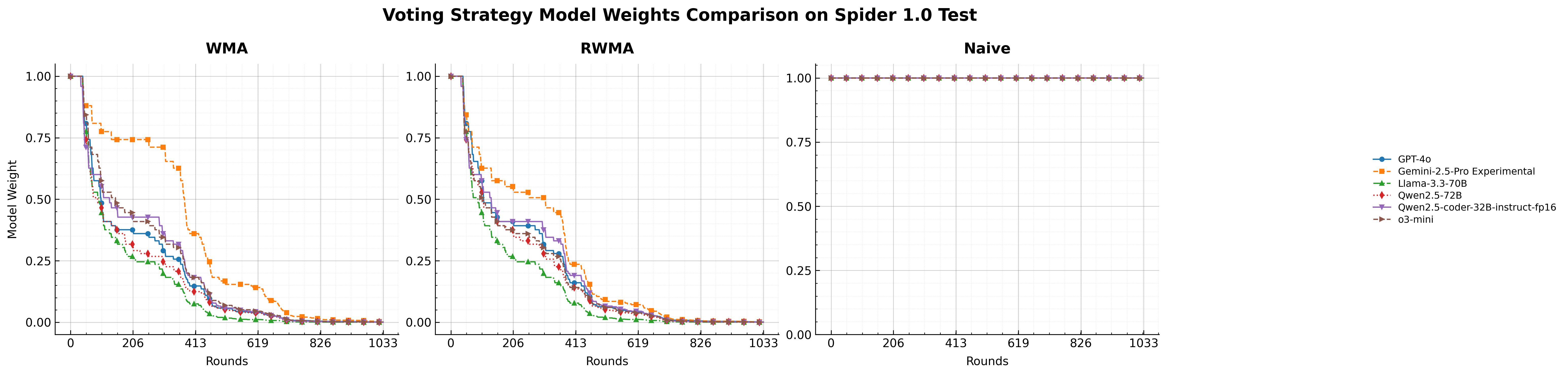}
\caption{Voting strategy model weights (Stage 6) on Spider 1.0 Test. WMA rapidly converges to stronger models.}
\label{fig:spider1_test_weights_best}
\end{figure}

\begin{table}[htb!]
\centering
\caption{Execution accuracy (Spider 1.0 Test) - voting applied at each stage}
\begin{tabular}{l|ccc|cccccc}
\toprule
Stage  & WMA & RWMA & Naïve & *Llama & *Qwen2.5-1 & *Qwen2.5-2 & GPT-4o & o3-mini & *Gemini-2.5 \\
\midrule
1 & 0.854 & 0.826 & 0.799 & 0.786 & 0.809 & 0.809 & 0.840 & 0.832 & 0.857 \\
2 & 0.846 & 0.830 & 0.773 & 0.755 & 0.809 & 0.809 & 0.840 & 0.832 & 0.846 \\
3 & 0.860 & 0.845 & 0.797 & 0.779 & 0.812 & 0.815 & 0.838 & 0.832 & 0.868 \\
4 & 0.856 & 0.828 & 0.773 & 0.761 & 0.820 & 0.815 & 0.838 & 0.823 & 0.865 \\
5 & 0.860 & 0.845 & 0.797 & 0.779 & 0.812 & 0.815 & 0.838 & 0.832 & 0.868 \\
6.1  & \textbf{0.864} & -- & -- & 0.779 & 0.822 & 0.820 & 0.842 & 0.831 & 0.867 \\
6.2  &  -- & 0.830 & -- & 0.787 & 0.820 & 0.819 & 0.847 & 0.829 & 0.850 \\
6.3  & -- & -- & 0.786 & 0.774 & 0.822 & 0.816 & 0.843 & 0.823 & 0.856 \\

\midrule
        \multicolumn{10}{p{15cm}}{\footnotesize
        \textit{Note:} \par 
        Llama: Llama3.3
        Qwen2.5-1:Qwen2.5-72B \par
        Qwen2.5-2: Qwen2.5-coder-32B-instruct-fp16 \par 
        Gemini-2.5:  Gemini-2.5-Pro Experimental \par
        Stage 6.1: with WMA Vote \par 
        Stage 6.2: with RWMA Vote \par
        Stage 6.3: with Naïve Vote
        } \\

\bottomrule
\end{tabular}
\label{tab:spider1_test_results}
\end{table}
Stage 6.1 (i.e., combining WMA voting, execution-guided self-refinement, and schema linking) achieves the highest execution accuracy of 86.4\% on the test set. This result underscores the strong synergy between dynamic ensemble voting and iterative refinement mechanisms. As shown in \autoref{fig:spider1_test_weights_best}, voting dynamics at Stage 6.1 on the test set, WMA again exhibits rapid convergence within the first 300 rounds, effectively concentrating weights on the strongest models, notably Gemini-2.5-Pro Experimental and GPT-4o. 

\begin{figure}[htb!]
\centering
\includegraphics[width=\textwidth]{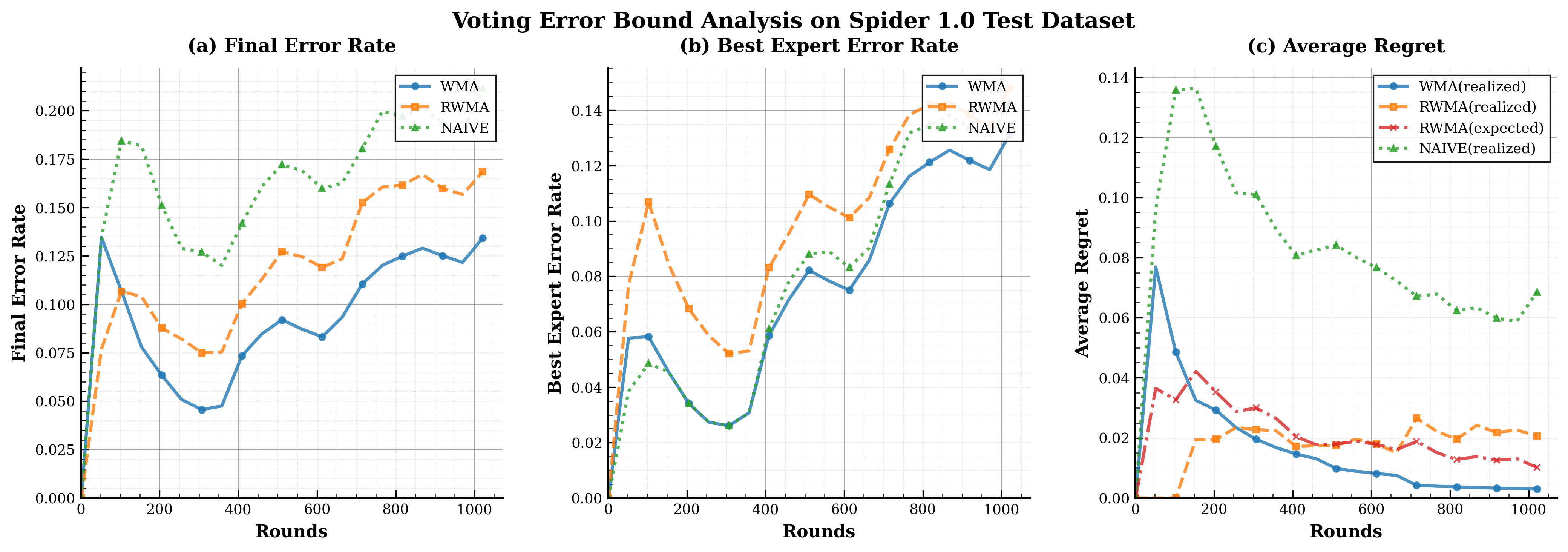}
\caption{Voting error bound analysis (Stage 6) on Spider 1.0 Test. WMA achieves the lowest final error rate and regret, almost approaching zero.}
\label{fig:spider1_test_error_best}
\end{figure}

\autoref{fig:spider1_test_error_best} shows that WMA consistently achieves the lowest final error rate and average regret. Similar to the dev dataset's results, the average regret of WMA more closely tracks the best expert error rate, further validating its effectiveness in dynamically adjusting to ensemble performance.

\subsubsection{Cross Stage Investigation on BIRD Dev}
\autoref{tab:bird_dev_results} reports execution accuracy on the BIRD development set across six incremental stages. Overall, we observe consistent performance improvements as schema linking, execution-guided self-refinement, and ensemble voting strategies are progressively incorporated. \autoref{tab:bird_dev_results} also show that introducing execution-guided self-refinement in Stage 3 yields a substantial performance boost across both individual models and the ensemble as a whole, indicating its effectiveness in correcting execution errors and improving robustness.

\begin{table}[htb!]
\centering
\caption{Execution accuracy (BIRD Dev) - voting applied at each stage.}
\begin{tabular}{l|ccc|cccccc}
\toprule
Stage  & WMA & RWMA & Naïve & *Llama & *Qwen2.5-1 & *Qwen2.5-2 & GPT-4o & o3-mini & *Gemini-2.5 \\
\midrule
1 & 65.51\% & 63.04\% & 62.39\% & 61.41\% & 57.56\% & 60.37\% & 64.02\% & 61.15\% & 65.97\% \\
2 & 66.10\% & 62.97\% & 62.26\% & 60.69\% & 57.69\% & 59.45\% & 63.17\% & 60.82\% & 66.23\% \\
3 & 66.04\% & 65.25\% & 62.78\% & 62.65\% & 59.65\% & 62.65\% & 64.54\% & 60.89\% & 66.82\% \\
4 & \textbf{66.30\%} & 64.15\% & 63.49\% & 63.17\% & 60.04\% & 62.58\% & 64.08\% & 61.21\% & 66.62\% \\
5 & \textbf{66.04\%} & 65.25\% & 62.78\% & 62.65\% & 59.65\% & 62.65\% & 64.54\% & 60.89\% & \textbf{66.82\%} \\
6.1  & \textbf{65.71\%} & -- & -- & 62.26\% & 60.37\% & 62.19\% & 64.93\% & 60.50\% & \textbf{66.56\%} \\
6.2  & -- & 64.15\% & -- & 62.19\% & 59.26\% & 62.26\% & 63.49\% & 61.21\% & 66.30\% \\
6.3   & -- & -- & 62.58\% & 61.86\% & 60.17\% & 62.45\% & 63.30\% & 60.37\% & 66.43\% \\
\midrule
        \multicolumn{10}{p{15cm}}{\footnotesize
        \textit{Note:} \par 
        Llama: Llama3.3
        Qwen2.5-1:Qwen2.5-72B \par
        Qwen2.5-2: Qwen2.5-coder-32B-instruct-fp16 \par 
        Gemini-2.5:  Gemini-2.5-Pro Experimental \par
        Stage 6.1: with WMA Vote \par 
        Stage 6.2: with RWMA Vote \par
        Stage 6.3: with Naïve Vote
        } \\
\bottomrule
\end{tabular}
\label{tab:bird_dev_results}
\end{table}

With the introduction of execution-guided self-refinement in Stage 3, ensemble accuracy under WMA voting increases to 66.04\%. The strongest overall performance is achieved in Stage 4, followed closely by Stage 6.1. Notably, Stage 6.1, which integrates WMA voting, execution-guided self-refinement, and schema linking, attains an execution accuracy of 65.71\%, approaching the performance of the best individual model, Gemini-2.5-Pro Experimental (66.56\%). This result highlights the effectiveness of combining refinement mechanisms with adaptive ensemble voting.

\begin{figure}[hbt!]
\centering
\includegraphics[width=\textwidth]{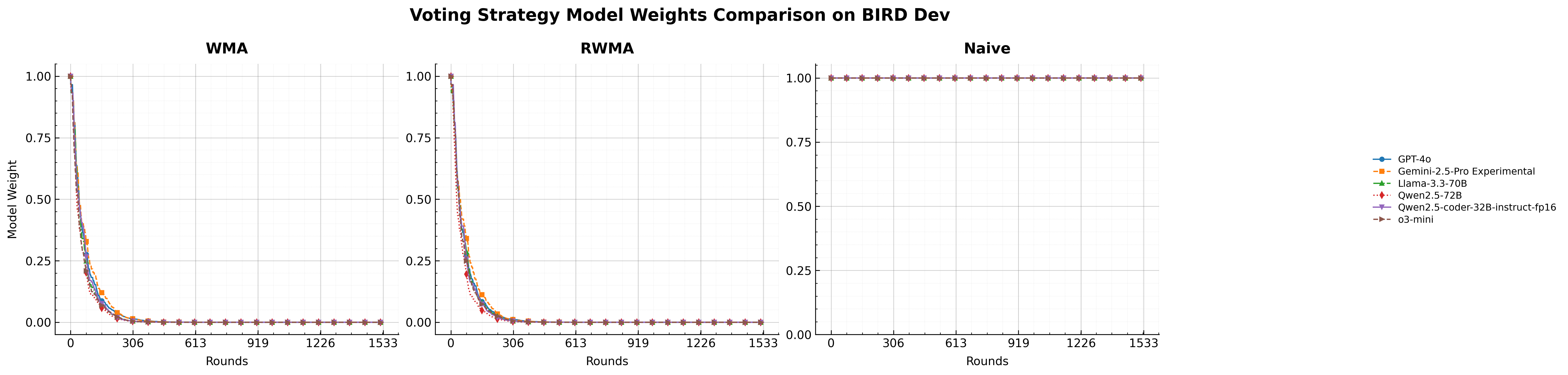}
\caption{Voting strategy model weights (Stage 6) on BIRD Dev. WMA converges quickly to stronger models.}
\label{fig:bird_dev_weights_best}
\end{figure}

\begin{figure}[hbt!]
\centering
\includegraphics[width=\textwidth]{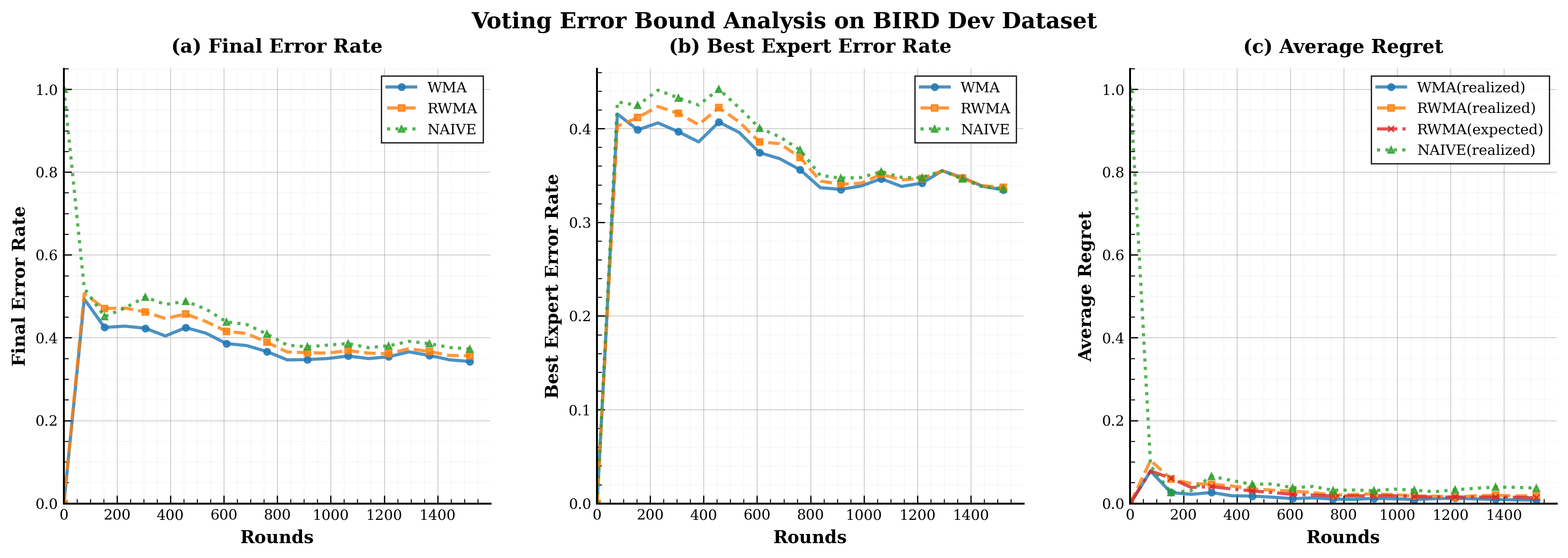}
\caption{Voting error bound analysis (Stage 6) on BIRD Dev. WMA achieves the lowest final error rate and regret.}
\label{fig:bird_dev_error_best}
\end{figure}

Figure~\ref{fig:bird_dev_weights_best} illustrates the voting dynamics at Stage 6.1 on the BIRD development set. Compared with the Spider 1.0 Dev and Test results, WMA on BIRD exhibits faster convergence, with model weights sharply differentiating within the first 300 rounds. This behavior reflects the more pronounced performance disparities among models on the BIRD benchmark. Despite rapid convergence under both WMA and RWMA, Gemini-2.5-Pro Experimental consistently dominates the weight distribution across rounds, a pattern also observed in the Spider 1.0 experiments. In contrast, Naïve voting maintains uniform weights throughout, lacking adaptability and consequently yielding inferior ensemble performance.

Figure~\ref{fig:bird_dev_error_best} further shows that WMA achieves the lowest final error rate and average regret among the three voting strategies, consistent with observations on Spider 1.0. This demonstrates WMA’s ability to adapt effectively toward expert-level performance over time. Compared with Spider 1.0, overall error rates on the BIRD development set are higher. The gap between WMA’s average regret and the best expert error rate is slightly larger on BIRD. Nevertheless, WMA remains the most robust and stable voting strategy, as its average regret consistently stays below that of RWMA and Naïve voting across all rounds. Moreover, WMA continues to adapt toward the strongest experts, demonstrating its effectiveness in handling domain variability and complex schema challenges inherent to the BIRD dataset.

\subsubsection{Investigation on Schema Linking}
Between stage 1 and 2, schema linking is the sole difference so their comparison constitutes a controlled ablation on schema linking, enabling us to isolate its impact. As shown in \autoref{tab:stage1_vs_2}, schema linking provides modest yet consistent performance gains, particularly for weaker models. 

\begin{table}[htb!]
    \centering
    \caption{Impact on Schema Linking (Stage 1 vs Stage 2) across datasets}
    \begin{tabular}{llcccccc}
        \toprule
        \textbf{Dataset} & \textbf{Stage} 
        & \textbf{Llama-3.3-70B} 
        & \textbf{Qwen2.5-72B} 
        & \textbf{*Qwen2.5-32B} 
        & \textbf{GPT-4o} 
        & \textbf{o3-mini} 
        & \textbf{*Gemini-2.5} \\
        \midrule
        \multirow{2}{*}{Spider 1.0 Dev} 
            & 1   & 77.3 & 79.0 & 82.3 & 79.9 & 81.7 & 84.9 \\
            & 2   & 77.6 & 80.2 & 82.6 & 81.0 & 81.7 & 84.8 \\
        \midrule
        \multirow{2}{*}{Spider 1.0 Test} 
            & 1  & 78.6 & 80.9 & 80.9 & 84.0 & 83.2 & 85.7 \\
            & 2   & 75.5 & 80.9 & 80.9 & 84.0 & 83.2 & 84.6 \\
        \midrule
        \multirow{2}{*}{BIRD} 
            & 1    & 61.41 & 57.56 & 60.37 & 64.02 & 61.15 & 65.97 \\
            & 2    & 60.69 & 57.69 & 59.45 & 63.17 & 60.82 & 66.23 \\
        \midrule
        \multicolumn{8}{p{15cm}}{\footnotesize
        \textit{Note:} \par 
        Unit in \% \par 
        Qwen2.5-32B: Qwen2.5-coder-32B-instruct-fp16 \par 
        Gemini-2.5:  Gemini-2.5-Pro Experimental
        } \\
        \bottomrule
    \end{tabular}
    \label{tab:stage1_vs_2}
\end{table}

\subsubsection{Investigation on Effectiveness of Execution-Guided Self-Refinement}
\autoref{tab:refinement_impact} highlights the effect of execution-guided self-refinement on Spider 1.0 Dev, Spider 1.0 Test, and BIRD Dev. Especially on BIRD Dev, execution-guided self-refinement yields the gains up to +2.28\% (Qwen2.5-coder-32B) and +2.21\%(RWMA). It demonstrates its robustness in handling complex, cross-domain queries or real-world scenarios. Conversely, Spider 1.0 Dev and Test show marginal or mixed effects, with most gains below +1.9\%. For the relatively strong model, Gemini-2.5-Pro-Experimental, even shows a slight loss and is less likely to benefit from the execution-guided self-refinement.

\begin{table}[htb!]
\centering
\caption{Impact on self-refinement (stage 1 vs stage 3) across datasets}
\begin{tabular}{l|ccc|ccc|ccc}
\toprule
{Model / Method} & \multicolumn{3}{c|}{Spider 1.0 Dev} & \multicolumn{3}{c|}{Spider 1.0 Test} & \multicolumn{3}{c}{BIRD Dev} \\
 Stage \& Gain   & 1 & 3 & Gain &  1 &  3 & Gain &  1 & 3 & Gain \\
\midrule
WMA               & \textbf{84.60} & 83.90 & \textbf{-0.70} & 85.40 & 86.00 & +0.60 & 65.51 & 66.04 & +0.53 \\
RWMA              & 81.40 & 82.80 & +1.40 & 82.60 & \textbf{84.50} & \textbf{+1.90} & 63.04 & 65.25 & \textbf{+2.21} \\
Naive             & 79.60 & 80.90 & +1.30 & 79.90 & 79.70 & \textbf{-0.20} & 62.39 & 62.78 & +0.39 \\
LLaMA3.3          & 77.30 & 78.40 & +1.10 & 78.60 & 77.90 & \textbf{-0.70} & 61.41 & 62.65 & +1.24 \\
Qwen2.5-72B       & 79.00 & 79.70 & +0.70 & 80.90 & 81.20 & +0.30 & 57.56 & 59.65 & +2.09 \\
Qwen2.5-coder-32B & 82.30 & 82.90 & +0.60 & 80.90 & 81.50 & +0.60 & 60.37 & \textbf{62.65} & \textbf{+2.28} \\
GPT-4o            & 79.90 & 79.90 &  0.00 & 84.00 & 83.80 & \textbf{-0.20} & 64.02 & 64.54 & +0.52 \\
o3-mini           & 81.70 & 81.70 &  0.00 & 83.20 & 83.20 &  0.00 & 61.15 & 60.89 & \textbf{-0.26} \\
Gemini2.5-Pro     & \textbf{84.90} & 84.10 & \textbf{-0.80} & \textbf{85.70} & 86.80 & +1.10 & \textbf{65.97} & 66.82 & +0.85 \\
\midrule
        \multicolumn{10}{p{15cm}}{\footnotesize \textit{Note:} Unit in \%} \\
\bottomrule
\end{tabular}
\label{tab:refinement_impact}
\end{table}

\subsubsection{Investigation on Effectiveness of Execution-Guided Self-Refinement and Schema Linking}
Table~\ref{tab:schema_refine_impact} demonstrates the impact of combining schema linking and execution-guided self-refinement (Stage 4) in comparison with the base model (Stage 1). Especially on the BIRD Dev set, moderate gains were observed across nearly three voting strategies and all models. Qwen2.5-coder-32B (+2.21\%) and Qwen2.5-72B (+2.48\%) achieve the highest improvements, highlighting not only the benefit of iterative refinement but also schema grounding in handling complex real-world queries. 
However, for Spider 1.0 Dev and Test, the performance shift was generally slight. We observe that some models (e.g., RWMA and GPT-4o on Dev) exhibited modest gains (+1.00\% and +1.40\% respectively), while others (e.g., Naïve and LLaMA3.3) slightly dropped. In summary, the combined effect is more impactful on structurally diverse benchmarks like BIRD, where schema complexity amplifies the utility of refinement signals. Improvements to Spider datasets are limited when the base execution quality is already high.

The key insight is that self-refinement is more impactful under higher uncertainty and complexity (e.g., BIRD Dev), especially when paired with RWMA(+2.21\%).

\begin{table}[htb!]
\centering
\caption{Impact on schema linking and self-refinement (stage 1 vs stage 4) across datasets.}
\begin{tabular}{l|ccc|ccc|ccc}
\toprule
{Model / Method} & \multicolumn{3}{c|}{Spider 1.0 Dev} & \multicolumn{3}{c|}{Spider 1.0 Test} & \multicolumn{3}{c}{BIRD Dev} \\
 Stage \& Gain   & 1 & 4 & Gain &  1 &  4 & Gain &  1 & 4 & Gain \\
\midrule
WMA               & 84.60 & 84.40 & -0.20 & 85.40 & 85.60 & +0.20 & 65.51 & 66.30 & +0.79 \\
RWMA              & 81.40 & 82.40 & +1.00 & 82.60 & 82.80 & +0.20 & 63.04 & 64.15 & +1.11 \\
Naive             & 79.60 & 79.50 & -0.10 & 79.90 & 77.30 & -2.60 & 62.39 & 63.49 & +1.10 \\
LLaMA3.3          & 77.30 & 77.00 & -0.30 & 78.60 & 76.10 & -2.50 & 61.41 & 63.17 & +1.76 \\
Qwen2.5-72B       & 79.00 & 79.90 & +0.90 & 80.90 & 82.00 & +1.10 & 57.56 & 60.04 & \textbf{+2.48} \\
Qwen2.5-coder-32B & 82.30 & 81.90 & -0.40 & 80.90 & 81.50 & +0.60 & 60.37 & 62.58 & \textbf{+2.21} \\
GPT-4o            & 79.90 & 81.30 & +1.40 & 84.00 & 83.80 & -0.20 & 64.02 & 64.08 & +0.06 \\
o3-mini           & 81.70 & 81.90 & +0.20 & 83.20 & 82.30 & -0.90 & 61.15 & 61.21 & +0.06 \\
Gemini2.5-Pro     & 84.90 & 84.80 & -0.10 & 85.70 & 86.50 & +0.80 & 65.97 & 66.62 & +0.65 \\
\midrule
        \multicolumn{10}{p{15cm}}{\footnotesize \textit{Note:} Unit in \%} \\
\bottomrule
\end{tabular}

\label{tab:schema_refine_impact}
\end{table}

\subsubsection{Investigation on Ensemble Voting}
Stage 6 shows moderate gains in accuracy over the baseline.  As shown in Table~\ref{tab:stage6_1_gains}, WMA yields the most reliable improvements (+0.90\% on Spider 1.0 Dev, +1.00\% on Spider Test, +0.20\% on BIRD), while RWMA also achieves small gains (+1.50\%, +0.40\%, +0.61\%). Although Naïve voting shows minimal or even negative gains, particularly on the Spider 1.0 Test, it still demonstrates the effectiveness of adaptive ensemble voting strategies. Furthermore, while Gemini-2.5-Pro-Experimental kept the top-performing individual model, ensemble voting methods leveraging diverse LLM predictions proved more effective overall.

\begin{table}[htb!]
\centering
\caption{Execution Accuracy Gains from Voting + Self-Refinement + Schema Linking (Stage 6 vs Stage 1) across datasets.}
\begin{tabular}{l|ccc|ccc|ccc}
\toprule
{Model / Method} & \multicolumn{3}{c|}{Spider 1.0 Dev} & \multicolumn{3}{c|}{Spider 1.0 Test} & \multicolumn{3}{c}{BIRD Dev} \\
Stage \& Gain     & 1 & 6.1/6.2/6.3 & Gain & 1 & 6.1/6.2/6.3 & Gain & 1 & 6.1/6.2/6.3 & Gain \\
\midrule
WMA              & 84.60 & 85.50 & +0.90 & 85.40 & 86.40 & +1.00 & 65.51 & 65.71 & +0.20 \\
RWMA               & 81.40 & 82.90 & +1.50 & 82.60 & 83.00 & +0.40 & 63.04 & 64.15 & +0.61 \\
Naive              & 79.60 & 79.70 & +0.10 & 79.90 & 78.60 & -1.30 & 62.39 & 62.58 & +0.17 \\
\midrule
        \multicolumn{10}{p{15cm}}{\footnotesize \textit{Note:} Unit in \%} \\
\bottomrule
\end{tabular}
\label{tab:stage6_1_gains}
\end{table}

\subsection{ReCAPAgent-SQL}
\label{sec:recap_spider2}
To evaluate the effectiveness of ReCAPAgent-SQL, we compare multiple system variants against strong baselines using the Spider 2.0-lite benchmark. Since the full Spider 2.0 dataset is still under active development during our study, our experiments are conducted on the first 100 queries of the Spider 2.0-lite benchmark.

We evaluate six experimental settings to analyze the impact of ReCAPAgent-SQL, model choice, and ensemble voting strategies:

\begin{itemize}
  \item \textbf{Setting 1 (Grok-3-beta)}: The official Spider-Agent implementation using the Grok-3-beta model. This setting serving as a baseline. 
  \item \textbf{Setting 2 (ReCAPAgent-SQL + Grok-3-beta)}: ReCAPAgent-SQL integrated with the Grok-3-beta model.
  \item \textbf{Setting 3 (GPT-4.1)}: The official Spider-Agent implementation using the GPT-4.1 model. This setting serving as another baseline. 
  \item \textbf{Setting 4 (ReCAPAgent-SQL + GPT-4.1)}: ReCAPAgent-SQL integrated with the GPT-4.1 model.
  \item \textbf{Setting 5 (ReCAPAgent-SQL + GPT-4.1 + WMA)}: ReCAPAgent-SQL with GPT-4.1 and the WMA for ensemble voting. 
  \item \textbf{Setting 6 (ReCAPAgent-SQL + GPT-4.1 + RWMA)}: ReCAPAgent-SQL with GPT-4.1 and the RWMA for ensemble voting. 
\end{itemize}

\subsubsection{Cross-Setting Analysis of ReCAPAgent-SQL}
Across all settings, ReCAPAgent-SQL demonstrates substantial performance gains over baseline systems. When applied to Grok-3-beta, ReCAPAgent-SQL improves execution accuracy from 23\% to 29\%, representing a 6\% absolute gain. The improvements are even more pronounced for GPT-4.1, where ReCAPAgent-SQL boosts execution accuracy from 6\% to 31\%, a 25\% absolute improvement.

Among ensemble strategies, WMA achieves the highest overall execution accuracy (31\%), slightly outperforming RWMA (29\%). However, performance on hard queries remains challenging: RWMA exhibits a modest advantage on hard instances, achieving 12.5\% accuracy compared to 8.3\% for WMA. These results suggest a trade-off between overall accuracy and robustness on harder queries.

\autoref{tab:spider2_recap} summarizes execution accuracy by query difficulty. Overall, the full ReCAPAgent-SQL configuration with WMA voting achieves the best total execution accuracy on Spider 2.0-lite, demonstrating the effectiveness of combining execution-guided refinement with adaptive ensemble voting.

\begin{table}[ht]
\centering
\caption{Execution accuracy on Spider 2.0-lite's first 100 queries by difficulty. Our full ReCAPAgent-SQL with WMA voting achieves the highest overall execution accuracy.}
\begin{tabular}{lcccc}
\toprule
\textbf{Setting} & \textbf{Easy} & \textbf{Medium} & \textbf{Hard} & \textbf{Total (100)} \\
\midrule
1. Grok-3-beta & 38.1\% & 20.0\% & 16.7\% & 23.0\% \\
2. ReCAPAgent-SQL with Grok-3-beta & 42.9\% & 29.1\% & 16.7\% & 29.0\% \\
3. GPT-4.1 & 19.0\% & 3.6\%  & 0.0\%  & 6.0\%  \\
4. ReCAPAgent-SQL with GPT-4.1 & 52.4\% & 32.7\% & 8.3\% & 31.0\% \\
5. ReCAPAgent-SQL with GPT-4.1 and WMA & 52.4\% & 32.7\% & 8.3\% & \textbf{31.0\%} \\
6. ReCAPAgent-SQL with GPT-4.1 and RWMA & 47.6\% & 29.1\% & \textbf{12.5\%} & 29.0\% \\
\bottomrule
\end{tabular}

\label{tab:spider2_recap}
\end{table}

\begin{figure}[ht]
\centering
\includegraphics[width=\textwidth]{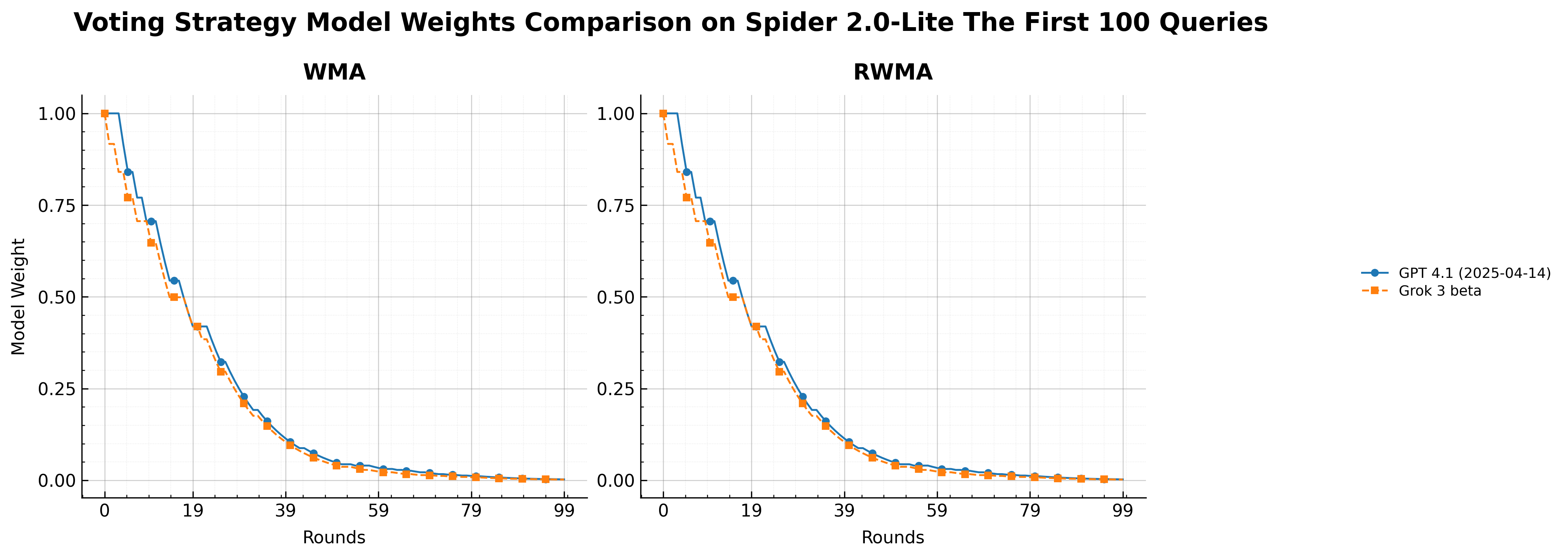}
\caption{Performance comparison of baseline and ReCAPAgent-SQL on 100 Spider 2.0-lite queries. GPT-4.1 shows significant improvements.}
\label{fig:recap_weight_effect}
\end{figure}

\begin{figure}[ht]
\centering
\includegraphics[width=0.85\textwidth]{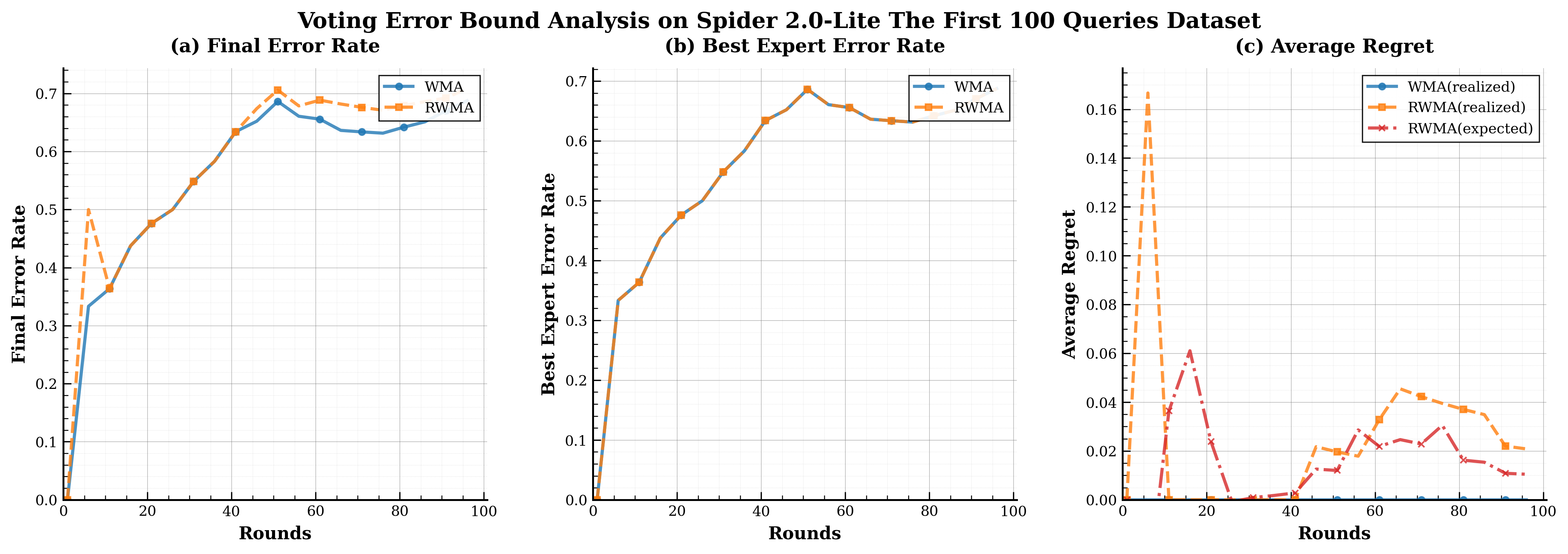} 
\caption{Ensemble Voting strategy analysis: WMA vs RWMA on Spider 2.0-lite. WMA achieves slightly higher overall accuracy.}
\label{fig:recap_voting_effect}
\end{figure}

\autoref{fig:recap_weight_effect} and \autoref{fig:recap_voting_effect} further illustrate the impact of ReCAPAgent-SQL and the voting strategies. Applying ReCAPAgent-SQL substantially improves performance for both Grok-3-beta and GPT-4.1. GPT-4.1 consistently outperforms Grok-3-beta under both WMA and RWMA strategies. Moreover, Grok-3-beta’s weight decreases slightly faster than GPT-4.1 over time, reflecting its relatively weaker performance across rounds.

\autoref{fig:recap_voting_effect} provides further insight into the ensemble voting dynamics. Panel (a) indicates that the final error rates of both WMA and RWMA increase over successive rounds; however, WMA consistently exhibits a lower error rate than RWMA during the final 50 rounds. Panel (c) shows that RWMA accumulates higher realized and expected regret, underscoring WMA’s stronger capacity to adaptively prioritize higher-performing experts throughout the ensemble process. Collectively, these observations substantiate the effectiveness and robustness of ReCAPAgent-SQL and its voting mechanisms on the Spider 2.0-lite benchmark.

\subsubsection{Voting Integration in a Multi-Agent Context}

The 25\% performance improvement after integrating ReCAPAgent-SQL with GPT-4.1 demonstrates its effectiveness in complex and realistic settings. This improvement arises from the coordinated interaction of multiple specialized agents, including structured reasoning via the PlannerAgent, iterative feedback through the CritiqueAgent, and adaptive schema correction enabled by the SchemaLinkerAgent. Notably, these agents are designed based on insights derived from our SSEV pipeline, and they facilitate robust behavior under execution errors, schema ambiguity, and incomplete or under-specified query conditions.

Within the ReCAPAgent-SQL framework, ensemble voting strategies (WMA and RWMA) are used to select final SQL outputs without access to ground-truth supervision during inference. Overall, WMA achieves higher total execution accuracy (31\%) than RWMA (28\%), while RWMA exhibits a slight advantage on hard queries. This trade-off highlights the complementary strengths of deterministic and randomized voting in complex Text-to-SQL scenarios.

Overall, our experimental results lead to the following conclusions. Execution-guided self-refinement combined with WMA substantially improves robustness, particularly under schema ambiguity. Schema linking further amplifies the benefits of refinement, especially in complex datasets. Among the evaluated ensemble strategies, WMA proves to be the most effective and stable, achieving expert-level accuracy with low regret. Finally, ReCAPAgent-SQL significantly enhances adaptability in realistic, evolving environments by supporting iterative, error-aware refinement.

These findings confirm the complementary roles of refinement, schema reasoning, and adaptive ensemble voting in advancing the robustness and effectiveness of modern Text-to-SQL systems.

\section{Conclusions}
This work demonstrates that both the SSEV pipeline and ReCAPAgent-SQL achieve strong and consistent performance across multiple benchmarks, including Spider 1.0 Dev, Spider 1.0 Test, BIRD Dev, and Spider 2.0-lite. The results show that integrating execution-guided self-refinement, schema linking, and ensemble voting substantially improves execution accuracy in Text-to-SQL tasks. In particular, the Weighted Majority Algorithm (WMA), when combined with refinement and schema-aware reasoning, yields performance that closely approaches the best individual expert. Moreover, incorporating ReCAPAgent-SQL on top of GPT-4.1 leads to significant accuracy gains, underscoring its effectiveness in complex, ambiguous, and realistic SQL generation scenarios. Overall, our findings emphasize the robustness of ensemble voting strategies such as WMA and highlight the strong potential of ReCAPAgent-SQL for handling high-complexity, real-world Text-to-SQL tasks.

In summary, this study introduces robust and theoretically grounded methods that demonstrate practical effectiveness on real-world-inspired benchmarks. By unifying adaptive refinement, schema-aware reasoning, and ensemble voting, our approach helps bridge the gap between academic Text-to-SQL research and enterprise-ready deployment, ultimately improving the reliability and accessibility of data-driven decision-making systems.

\section{Future Work}
Several promising directions remain for future exploration to enhance the accuracy, scalability, and enterprise applicability of ReCAPAgent-SQL. Future evaluations will extend beyond the first 100 queries of Spider 2.0-lite to the full Spider 2.0, Spider 2.0-lite, and Spider 2.0-snow benchmarks, capturing diverse and complex enterprise workflows to provide a more comprehensive assessment of robustness, generalization, and production readiness. Concurrently, the multi-agent architecture can be refined using open protocols such as Model Context Protocol (MCP) and Agent-to-Agent (A2A), enabling standardized, secure, and event-driven communication among agents, improving modularity, reducing format inconsistencies, and simplifying debugging. Finally, the SelfRefinerAgent can be augmented with retrieval-augmented generation (RAG), leveraging historical self-refinement logs and corrected SQL examples as few-shot guidance for subsequent iterations, reducing repeated errors and redundant exploration while further enhancing efficiency and accuracy in complex query scenarios.

\bibliographystyle{unsrt}  
\bibliography{references.bib}

@article{Xue2023DBGPT,
  author  = {Siqiao Xue and Caigao Jiang and Wenhui Shi and Fangyin Cheng and Keting Chen and Hongjun Yang and Zhiping Zhang and Jianshan He and Hongyang Zhang and Ganglin Wei and Wang Zhao and Fan Zhou and Danrui Qi and Hong Yi and Shaodong Liu and Faqiang Chen},
  title   = {DB-GPT: Empowering Database Interactions with Private Large Language Models},
  journal = {arXiv},
  year    = {2023},
  url     = {https://arxiv.org/abs/2312.17449},
}

@article{Hong2024NextGen,
  author  = {Zijin Hong and Zheng Yuan and Qinggang Zhang and Hao Chen and Junnan Dong and Feiran Huang and Xiao Huang},
  title   = {Next-Generation Database Interfaces: A Survey of LLM-based Text-to-SQL},
  journal = {arXiv},
  year    = {2024},
  url     = {https://arxiv.org/abs/2406.08426},
}

@misc{WrenAI2024GenBI,
  author = {{Wren AI}},
  title = {Revolutionize Business Intelligence with GenBI},
  year = {2024},
  url = {https://getwren.ai/genbi},
  note = {Accessed: 2025-05-21}
}

@inproceedings{yu-2018-spider,
  author = {Yu, Tao and Zhang, Rui and Yang, Kai and Yasunaga, Michihiro and Wang, Dongxu and Li, Zifan and Ma, James and Yao, Quan and Roman, Dragomir Radev},
  title = {Spider: A Large-Scale Human-Labeled Dataset for Complex and Cross-Domain Semantic Parsing and Text-to-SQL Task},
  booktitle = {Proceedings of the 2018 Conference on Empirical Methods in Natural Language Processing (EMNLP)},
  year = {2018},
  pages = {3911--3921},
  publisher = {Association for Computational Linguistics},
  url = {https://aclanthology.org/D18-1425},
}

@article{li2023bird,
  title   = {Can LLM Already Serve as A Database Interface? A BIg Bench for Large-Scale Database Grounded Text-to-SQLs},
  author  = {Jinyang Li and Binyuan Hui and Ge Qu and Jiaxi Yang and Binhua Li and Bowen Li and Bailin Wang and Bowen Qin and Rongyu Cao and Ruiying Geng and Nan Huo and Xuanhe Zhou and Chenhao Ma and Guoliang Li and Kevin C.C. Chang and Fei Huang and Reynold Cheng and Yongbin Li},
  journal = {arXiv preprint arXiv:2305.03111},
  year    = {2023},
  url     = {https://arxiv.org/abs/2305.03111}
}

@inproceedings{lei-2024,
  author = {Lei, Fangyu and Chen, Jixuan and Ye, Yuxiao and Cao, Ruisheng and Shin, Dongchan and Su, Hongjin and Suo, Zhaoqing and Gao, Hongcheng and Hu, Wenjing and Yin, Pengcheng and Zhong, Victor and Xiong, Caiming and Sun, Ruoxi and Liu, Qian and Wang, Sida and Yu, Tao},
  title = {Spider 2.0: Evaluating Language Models on Real-World Enterprise Text-to-SQL Workflows},
  booktitle = {Proceedings of the 12th International Conference on Learning Representations (ICLR)},
  year = {2025},
  month = may,
  note = {Oral Presentation},
  url = {https://arxiv.org/abs/2411.07763}
}

@article{Pourreza2023din,
  title   = {DIN-SQL: Decomposed In-Context Learning of Text-to-SQL with Self-Correction},
  author  = {Mohammadreza Pourreza, Davood Rafiei},
  journal = {arXiv preprint arXiv:2304.11015},
  year    = {2023},
  url     = {https://arxiv.org/abs/2304.11015}
}

@article{deng2025reforce,
  title   = {{ReFoRCE: A Text-to-SQL Agent with Self-Refinement, Consensus Enforcement, and Column Exploration}},
  author  = {Minghang Deng and Ashwin Ramachandran and Canwen Xu and Lanxiang Hu and Zhewei Yao and Anupam Datta and Hao Zhang},
  journal = {arXiv preprint arXiv:2502.00675},
  year    = {2025},
  url     = {https://arxiv.org/abs/2502.00675}
}

@article{li2024pet,
  title   = {{PET-SQL: A Prompt-Enhanced Two-Round Refinement of Text-to-SQL with Cross-consistency}},
  author  = {Zhishuai Li and Xiang Wang and Jingjing Zhao and Sun Yang and Guoqing Du and Xiaoru Hu and Bin Zhang and Yuxiao Ye and Ziyue Li and Rui Zhao and Hangyu Mao},
  journal = {arXiv preprint arXiv:2403.09732},
  year    = {2024},
  url     = {https://arxiv.org/abs/2403.09732}
}

@book{hazan2023oco,
  title     = {Introduction to Online Convex Optimization (Second Edition)},
  author    = {Elad Hazan},
  year      = {2023},
  note      = {Available at \url{https://arxiv.org/abs/1909.05207}},
  publisher = {arXiv preprint arXiv:1909.05207}
}

@misc{shalevCMU15850,
  author       = {Shai Shalev-Shwartz},
  title        = {Lecture 15: The Weighted Majority Algorithm},
  howpublished = {\url{https://www.cs.cmu.edu/~15850/notes/lec15.pdf}},
  note         = {Accessed: 2025-05-22},
  year         = {2022},
  institution  = {Carnegie Mellon University},
}

@article{gao2023texttosql,
  title   = {Text-to-SQL Empowered by Large Language Models: A Benchmark Evaluation},
  author  = {Dawei Gao and Haibin Wang and Yaliang Li and Xiuyu Sun and Yichen Qian and Bolin Ding and Jingren Zhou},
  journal = {arXiv preprint arXiv:2308.15363},
  year    = {2023},
  url     = {https://arxiv.org/abs/2308.15363}
}

@article{grattafiori2024llama3,
  title     = {The LLaMA 3 Herd of Models},
  author    = {Aaron Grattafiori and Abhimanyu Dubey and Abhinav Jauhri and others},
  journal   = {arXiv preprint arXiv:2407.21783},
  year      = {2024},
  url       = {https://arxiv.org/abs/2407.21783},
  note      = {Accessed: 2025-06}
}

@misc{deepmind2025gemini2.5,
  author       = {Google DeepMind},
  title        = {Gemini 2.5: Our Most Intelligent AI Model},
  year         = {2025},
  howpublished = {\url{https://deepmind.google/technologies/gemini}},
  note         = {Accessed: 2025-06}
}

@misc{xai2025grok3,
  author       = {xAI},
  title        = {Grok 3 Beta — The Age of Reasoning Agents},
  year         = {2025},
  howpublished = {\url{https://x.ai/news/grok-3}},
  note         = {Accessed: 2025-06}
}

@misc{qwen2024hf,
    title = {Qwen2.5: A Party of Foundation Models},
    url = {https://qwenlm.github.io/blog/qwen2.5/},
    author = {Qwen Team},
    month = {September},
    year = {2024}
}

@article{hui2024qwen2_5coder,
  title     = {Qwen2.5‑Coder Technical Report},
  author    = {Hui, Binyuan and Yang, Jian and Cui, Zeyu and others},
  journal   = {arXiv preprint arXiv:2409.12186},
  year      = {2024},
  url       = {https://arxiv.org/abs/2409.12186},
  note      = {Accessed: 2025-06}
}

@misc{openai2024gpt4o,
  author       = {OpenAI},
  title        = {GPT-4o: Fast, intelligent, flexible GPT model},
  year         = {2024},
  howpublished = {\url{https://platform.openai.com/docs/models/gpt-4o}},
  note         = {Accessed: 2025-06}
}

@misc{openai2025o3mini,
  author       = {OpenAI},
  title        = {o3-mini - A small model alternative to o3},
  year         = {2025},
  howpublished = {\url{https://platform.openai.com/docs/models/o3-mini}},
  note         = {Accessed: 2025-06}
}

@misc{openai2025gpt41,
  author       = {OpenAI},
  title        = {GPT-4.1 - Flagship GPT model for complex tasks},
  year         = {2025},
  howpublished = {\url{https://platform.openai.com/docs/models/gpt-4.1}},
  note         = {Accessed: 2025-04-23}
}

@misc{salesforce_text2sql,
  author       = {Salesforce},
  title        = {How We Built a Text-To-SQL AI Agent to Get Instant Answers From Our Data},
  howpublished = {\url{https://www.salesforce.com/blog/text-to-sql-agent/}},
  year         = {2025},
  note         = {Accessed: 2025-06}
}

@misc{snowflake_aisql,
  author       = {Snowflake},
  title        = {Snowflake Introduces Cortex AISQL and SnowConvert AI: Analytics Rebuilt for the AI Era},
  howpublished = {\url{https://www.snowflake.com/en/news/press-releases/snowflake-introduces-cortex-aisql-and-snowconvert-ai-analytics-rebuilt-for-the-ai-era/}},
  year         = {2025},
  note         = {Accessed: 2025-06}
}

\clearpage
\appendix
\section*{Appendix}
\textbf{A. Self-Refinement Prompt}
\vspace{0.25em}
\hrule  
\begin{verbatim}
-- Target SQL Dialect: BIGQUERY
[Project Information]
Current Project ID: bigquery-public-data
Important: Use these IDs in your queries.
...

[Database Context]
Current Database: ga4

[Database Verification Steps]
1. First, check the database structure:
	- Use 'ls' to see all files and directories
	- Look for files ending in .sql, .ddl, or .schema
	- Check for any README.md or documentation files
...

[Original SQL]
WITH EngagementData AS (
	SELECT
		user_pseudo_id, event_timestamp, (SELECT value.int_value FROM UNNEST(event_params) 
        WHERE key = 'engagement_time_msec') AS engagement_time_msec 
	FROM
		bigquery-public-data.ga4_obfuscated_sample_ecommerce.events_*
	WHERE
		_TABLE_SUFFIX BETWEEN '20210101' AND '20210107'
),
...
)
SELECT
	COUNT(DISTINCT user_pseudo_id) AS distinct_pseudo_users
FROM
	UsersWithEngagement7Days
WHERE
	user_pseudo_id NOT IN (SELECT user_pseudo_id FROM UsersWithEngagement2Days)

[Previous Execution Result]
Error occurred while fetching data: 400 No matching signature for operator BETWEEN
	Argument types: INT64, TIMESTAMP, TIMESTAMP
	Signature: (T1) BETWEEN (T1) AND (T1)
	Unable to find common supertype for templated argument <T1>
	Input types for <T1>: {INT64, TIMESTAMP} at; reason: invalidQuery,
	location: query, message: No matching signature for operator BETWEEN
	Location: US
	Job ID: ad6ed261-dc0b-494d-8187-e455ca555aa8
An unknown error occurred. Please re-check the SQL syntax and 
verify database schema alignment.

[Detected Error Type:] OtherError
Select one of the following strategies to apply:
1. Try simplifying query structure.
2. Focus on filtering conditions.
3. Try SELECT with minimal columns first.
4. Double-check all referenced schema components.
Explain your strategy choice and apply it to refine the SQL.

[Expected Output Format]
CSV Format: distinct_pseudo_users_count (integer)
Ensure the output matches this format exactly.

[Next Steps]
1. First verify the database structure and table names
2. If directory access fails, try alternative paths
3. Document any access issues encountered
4. Then refine the SQL query to resolve the issue
5. Ensure the output format matches the requirements

[Required Action Format]
You must output exactly one of the following actions (no other text):
- Action: BIGQUERY_EXEC_SQL(sql_query="...", is_save=..., save_path=".../result.csv")
- Action: Terminate(output=".../result.csv")

Important: When dealing with numerical results, 
DO NOT round the numbers. 
Keep the full precision of the results.
\end{verbatim}
\vspace{0.2em}
\hrule

\vspace{5em}

\textbf{B. Predict Action Prompt}
\vspace{0.25em}
\hrule  
\begin{verbatim}
[User Question]
How many distinct pseudo users had positive engagement time in the 7-day period ending 
on January 7, 2021 at 23:59:59, but had no positive engagement time in the 2-day period 
ending on the same date (January 7, 2021 at 23:59:59)?


[Project Information]
Current Project ID: bigquery-public-data
Important: Use these IDs in your queries.
Example: `project_id.dataset_id.table_id`
For public datasets, you can use: `bigquery-public-data.dataset_id.table_id`

[Database Context]
Current Database: ga4

[Database Verification Steps]
1. First, check the database structure:
   - Use 'ls' to see all files and directories
   - Look for files ending in '.sql', '.ddl', or '.schema'
   - Check for any README.md or documentation files

2. Verify table names and schema:
   - Open any `.sql` or `.ddl` files to see table definitions
   - Look for `CREATE TABLE` statements
   - Note the exact database and schema names
   - Check for any table aliases or views
   - If schema file not found, try other directories
......

[Current Plan]
['Identify the relevant table in the `ga4` dataset within the bigquery-public-data project 
that contains user engagement data, focusing on columns related to user identifiers and 
engagement time metrics, such as user_pseudo_id and engagement_time_msec', ...]

[Execution Result]

You are in the folder now.

[Reference Syntax]

# BigQuery Syntax Reference
## I need clarification and documentation on the following BigQuery-specific syntax and 
functions to ensure accurate implementation of the query:
1. **TIMESTAMP Functions and Timezone Handling**: I need documentation or examples on 
how to handle timestamps and timezones in BigQuery

"description": "GoogleSQL for BigQuery supports the following timestamp functions.\n\n
IMPORTANT: Before working with these functions", ......,
```

## specifically for filtering data within defined date ranges (e.g.

DATE | string | SQL input: ` DATE '2017-03-06' `
## January 1

`DATE`: {
## 2021 00:00:00 to January 7

`DATE`: {
## 2021 23:59:59). For instance

[Required Action Format]
You must output exactly one of the following actions (no other text):
Important: When dealing with numerical results, DO NOT round the numbers. 
Keep the full precision of the results.

- Action: BIGQUERY_EXEC_SQL(sql_query="...", is_save=..., save_path=".../result.csv")
- Action: Terminate(output=".../result.csv")
- The result matches the expected format
- The data has been properly saved to result.csv

Generate your next action.

\end{verbatim}
\vspace{0.2em}
\hrule

\end{document}